\documentclass{article} %

\usepackage{iclr2026_conference}

\usepackage{amsmath,amsfonts,bm}

\def\eqref#1{equation~\ref{#1}}

\def\1{\bm{1}}

\DeclareMathAlphabet{\mathsfit}{\encodingdefault}{\sfdefault}{m}{sl}
\SetMathAlphabet{\mathsfit}{bold}{\encodingdefault}{\sfdefault}{bx}{n}

\newcommand{\R}{\mathbb{R}}

\usepackage{hyperref}
\usepackage{url}
\usepackage{booktabs}       %
\usepackage{amsfonts}       %
\usepackage{nicefrac}       %
\usepackage{microtype}      %
\usepackage{xcolor}         %
\usepackage{amsmath}
\usepackage{amssymb}        %
\usepackage{float}          %
\usepackage{algorithm}      %
\usepackage{algpseudocode}  %
\usepackage{enumitem}
\usepackage{graphicx}

\title{Inference-Time Compute Scaling for Flow Matching}

\author{\textbf{Adam Stecklov}$^{1,2}$\thanks{Correspondence to: \url{adam.stecklov@mail.mcgill.ca} \\ \parindent 1.75em\indent Code available at: \url{https://github.com/adamkutak/tree-flow-matching}} \quad \textbf{Noah El Rimawi-Fine}$^{1,2}$ \quad \textbf{Mathieu Blanchette}$^{1,2}$
\\
\\
$^1$McGill University, $^2$Mila – Quebec AI Institute
}

\begin{document}

\maketitle

\begin{abstract}
Allocating extra computation at inference time has recently improved sample quality in large language models and diffusion-based image generation. In parallel, Flow Matching (FM) has gained traction in language, vision, and scientific domains, but inference-time scaling methods for it remain under-explored. Concurrently, Kim et al., 2025 approach this problem but replace the linear interpolant with a non-linear variance-preserving (VP) interpolant at inference, sacrificing FM's efficient and straight sampling. Additionally, inference-time compute scaling for flow matching has only been applied to visual tasks, like image generation. We introduce novel inference-time scaling procedures for FM that preserve the linear interpolant during sampling. Evaluations of our method on image generation, and for the first time (to the best of our knowledge), unconditional protein generation, show that I) sample quality consistently improves as inference compute increases, and II) flow matching inference-time scaling can be applied to scientific domains.
\end{abstract}

\begin{figure}[H]
  \centering
  \includegraphics[width=0.9\textwidth]{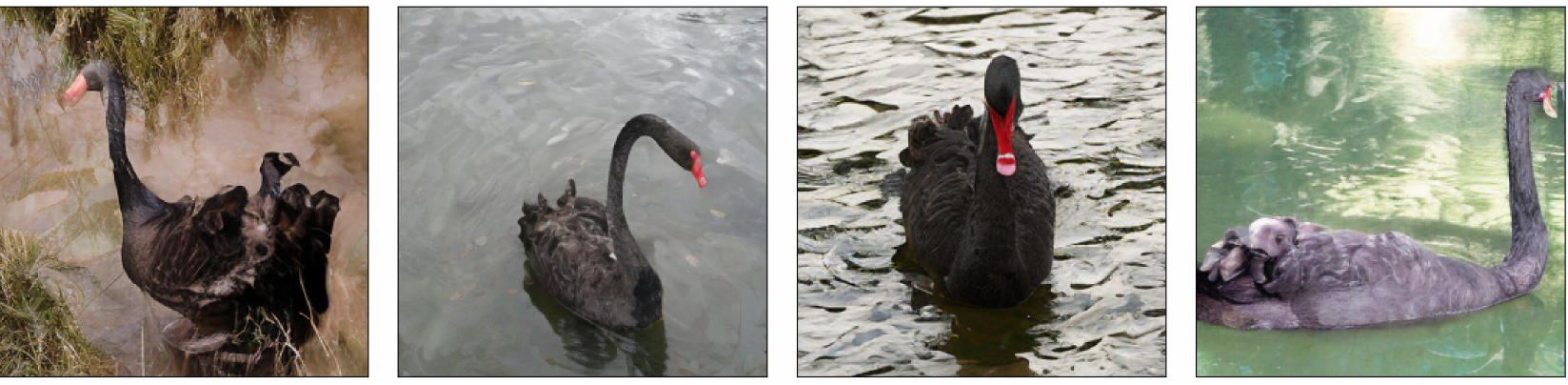}
  \caption{Sample quality improvement with inference-time compute scaling using our RS+NS–DMFM-ODE method on ImageNet 256×256 with the DINO verifier. From left to right: 1×, 2×, 4×, 8× compute budget. Higher compute budgets produce more coherent and detailed images.}
  \label{fig:dino-two-stage-samples}
\end{figure}

\section{Introduction}

Inference-time compute scaling improves generative models by spending more test-time computation without retraining. In discrete settings this includes longer reasoning chains, planning, and verifier-guided selection~\cite{gandhi2024sos, cobbe2021verifiers, lightman2023verify, brown2024llmmonkeys}, with notable examples such as OpenAI's o1 and o3 models ~\cite{openai2024o1} and DeepSeek R1~\cite{deepseek2025r1}. In the continous setting, verifier-guided search for diffusion models ~\cite{ma2025diffits} has been introduced and demonstrates improvement in image generation quality as inference-time increases.

Flow Matching ~\cite{lipman2023fm} trains a neural velocity field by linearly interpolating between the data distribution and the initial distribution (which, unlike diffusion, is not restricted to a gaussian), and samples straight deterministic paths with an ODE in fewer sampling steps than diffusion. it has achieved state of the art performance in image generation~\cite{flux}, and most notably in scientific domains such as protein folding and structure design~\cite{foldflow2}, small molecule generation ~\cite{flowmol3}, and faster Molecular Dynamics simulations ~\cite{nam2025flowmatchingacceleratedsimulation}.

However, the diffusion inference-time scaling technique introduced in ~\cite{ma2025diffits}  is not general to FMs formulation, which does not assume gaussian prior distributions. ~\cite{kim2025flowits} addresses this by introducing an inference-time scaling algorithm for FM, which relies on transforming the linear interpolant, at inference, into a non-linear variance-preserving (VP) interpolant, effectively transforming it into a diffusion model. This enables diverse trajectories (their primary motivation) at the expense of FM's fast straight sampling paths. Both methods have only been empirically validated on image generation, yet flow and diffusion models have become popular in the scientific domain. Further, we believe inference-time compute scaling is better suited for scientific challenges, such as drug design or protein folding, where the downstream benefits of higher quality samples is far greater, and we are therefore willing to spend more compute time on these tasks.
\textbf{We are therefore jointly motivated to investigate inference-time compute scaling for flow matching, maintaining the linear interpolant, and to demonstrate its generalizability to tasks in the scientific domain.} Our core contributions are as follows:
\begin{itemize}[noitemsep,topsep=0pt,parsep=0pt,partopsep=0pt,leftmargin=*]
    \item \textbf{We present the first inference-time scaling technique for Flow Matching that preserves the linear interpolant at inference}, using the Noise Search algorithm, which achieves substantial improvements for relatively small increases in computation, and verify results on Imagenet256 image generation. Further, we present a two-stage method that first runs Best-of-N, then applies our Noise Search algorithm, taking advantage of FM's prior distribution invariance, and balancing exploration and exploitation to achieve state of the art results.
    \item We show the generalizability of our approach by demonstrating \textbf{the first application of inference-time scaling for flow matching to a scientific domain, specifically unconditional protein design using FoldFlow2} ~\cite{foldflow2}, demonstrating significant increases in protein designability as we scale compute.
    \item We \textbf{investigate noise schedules that push the sample diversity vs. quality pareto front outwards} compared to a traditional SDE. However we find that additional diversity does not necessarily improve search-scaling performance.
\end{itemize}

\section{Preliminaries}
\label{sec:preliminaries}

\paragraph{Notation.} Random variables are uppercase (e.g., $X$), and realizations are lowercase (e.g., $x$). Distributions at time $t\in[0,1]$ are denoted $\pi_t$ with density $p_t$; endpoints are $\pi_0=\pi_{\mathrm{ref}}$ and $\pi_1=\pi_{\mathrm{data}}$. Vector fields $v:\R^d\times[0,1]\to\R^d$ are time‑dependent and (unless stated otherwise) Lipschitz in $x$ and measurable in $t$. The continuity equation is
\begin{equation}
\label{eq:continuity}
\partial_t p_t(x) + \nabla\!\cdot\!\big(p_t(x)\,v(x,t)\big)=0,\qquad p_0,p_1\ \text{given}.
\end{equation}

\subsection{Flow Matching}

Flow Matching (FM)~\cite{lipman2023fm} defines a continuous bridge between a reference distribution \(\pi_{\mathrm{ref}}\), typically (but not necessarily) a Gaussian distribution, and a data distribution \(\pi_{\mathrm{data}}\), by modeling trajectories \(x_t\) that interpolate between samples \(x_0 \sim \pi_{\mathrm{ref}}\) and \(x_1 \sim \pi_{\mathrm{data}}\). A common choice is the linear path $x_t = (1-t)x_0 + tx_1$, though other conditional paths are possible and lead to generalized or conditional flow matching \cite{tong2024otfm, song2021score, albergo2023stochastic_interpolants}. This is done via a learned velocity field \(v_\theta(x, t)\) that satisfies the probability-flow ODE:
\[
\frac{dx_t}{dt} = v_\theta(x_t, t), \quad t \in [0,1].
\]
To train \(v_\theta\), a supervised loss is used where the ground truth velocity is known analytically:
\[
v^\star(x_t, t) = x_1 - x_0.
\]
This target arises because \(\frac{dx_t}{dt} = x_1 - x_0\) under linear interpolation. The training loss is then
\[
\mathcal{L}_{\mathrm{FM}} = \mathbb{E}_{x_0 \sim \pi_{\mathrm{ref}}, x_1 \sim \pi_{\mathrm{data}}, t \sim \mathcal{U}[0,1]} \left[ \left\| v_\theta(x_t, t) - (x_1 - x_0) \right\|^2 \right].
\]
While FM defines its objective using i.i.d. sample pairs \((x_0, x_1)\), such pairs are typically poorly coupled in high-dimensional space. Minibatch Optimal Transport Flow Matching (OT-FM)~\cite{tong2024otfm} addresses this by using an optimal transport plan computed within each minibatch to generate more meaningful pairs, shortening transport paths and improving training stability. \textbf{Critically, because the interpolant is linear and the learned dynamics are smooth, FM enables fewer sampling steps while maintaining sample quality. While flow matching can inherently be scaled at inference by reducing the stepsize, gains from this are well known to plateau quickly ~\cite{lipman2023fm}, motivating the use of novel scaling axes.}

\subsection{Stochastic Interpolants: A Unifying Framework}

The stochastic interpolants framework~\cite{albergo2023stochastic_interpolants, ma2024sit} shows that both FM and diffusion models can be described within the same interpolant framework. A stochastic interpolant is defined by
\[
x_t = a(t) x_0 + b(t) x_1 + \sigma(t) \epsilon, \quad \epsilon \sim \mathcal{N}(0, I),
\]
with boundary conditions $a(0)=1,\ b(0)=0,\ \sigma(0)=0$ and $a(1)=0,\ b(1)=1,\ \sigma(1)=0$,
and with $a,b,\sigma$ smooth in $t$. FM is the deterministic special case $a(t)=1-t$, $b(t)=t$, $\sigma(t)\equiv 0$.

Crucially, the stochastic interpolants framework allows inference-time reinterpretation of trained
models by simply modifying the schedule $(a,b,\sigma)$. For instance, a model trained with linear FM
($a(t)=1-t,\ b(t)=t,\ \sigma(t)\equiv 0$) can, at test time, be sampled using a different valid
schedule. One example is the variance-preserving (VP) diffusion schedule, usually written in the two-coefficient
form $x_t=\alpha_t x_{\mathrm{data}}+\sigma_t \epsilon$ with $\alpha_t^2+\sigma_t^2=1$.
This is a special case of the three-coefficient interpolant by reinterpreting $(x_0,x_1)=(\epsilon,
x_{\mathrm{data}})$ and setting $a(t)=\sigma_t,\ b(t)=\alpha_t,\ \sigma(t)=0$.
In both cases, the learned parameters can be reused without retraining, revealing a continuum of
models and motivating inference-time strategies that use the same learned velocity and score fields.

\section{Related Work}
\label{sec:related}

\subsection{Inference-Time Compute Scaling for Stochastic Interpolants models}
\label{sec:related-fmscaling}

~\cite{ma2025diffits} proposes 3 algorithms to enable inference-time scaling for diffusion models, where a verifier function \( r(\cdot) \) provides a score to completed samples and we aim to maximize this score. The first, Random Search (RS), proposes to treat the initial noise vector \( z \sim \mathcal{N}(0, I) \) as a controllable input, performing Best-of-N style sampling, i.e take the top-$K$ of $N$ generated samples. They then improve on this by introducing a first-order search algorithm which perturbs the top initial noises before resampling, taking advantage of prior information when searching for optimal initial noise vectors. Finally, they propose a \textit{search over paths} algorithm, which performs a limited example of exploring denoising paths, by iteratively de-noising, selecting, and re-noising the best candidates. In practice, search begins at \( t = 0.11 \) (where $t=0$ is a de-noised sample, as diffusion generally uses the inverse time notation of FM), and each re-noising step applies \( t' = 0.89 \) time worth of noise back to the sample (i.e 89\% of the noise is re-applied). This method therefore loses almost all the prior information signal from the samples at \( t \ge 0.11 \), effectively acting as a best-of-\( N \) method and failing to truly search the space of possible branching trajectories. Further, these algorithms depend on the prior distribution being gaussian, making them less generalizable and not applicable to the general flow matching case.

A concurrent line of work proposes inference-time scaling for flow models by introducing stochasticity and path diversity into the otherwise deterministic sampling process~\cite{kim2025flowits}. This is achieved through a two-step transformation of the learned model, leveraging the stochastic interpolants framework ~\cite{albergo2023stochastic_interpolants} to, at inference, (1) convert the velocity field from the probability flow ODE to a score-based SDE sampler and (2) convert the linear interpolant into the variance-preserving (VP) interpolant (most commonly used in diffusion models).

Specifically, the learned velocity \( u_t(x) \) is used to define the drift term of a reverse-time SDE:
\[
d x_t = f_t(x_t)\,dt + g_t\,dW_t, \quad \text{where } f_t(x_t) = u_t(x_t) - \frac{g_t^2}{2} \nabla \log p_t(x_t).
\]
The score function \( \nabla \log p_t(x_t) \) is estimated analytically from \( u_t \):
\[
\nabla \log p_t(x_t) = \frac{1}{\sigma_t} \cdot \frac{\alpha_t u_t(x_t) - \dot{\alpha}_t x_t}{\dot{\alpha}_t \sigma_t - \alpha_t \dot{\sigma}_t}.
\]
In parallel, the interpolation path is converted from a linear interpolant \( x_t = (1 - t)x_0 + t x_1 \) to a VP interpolant, such as \( x_t = \alpha_t x_0 + \sigma_t x_1 \) where \(\alpha_t^2 + \sigma_t^2 = 1\). This conversion requires transforming the original velocity field into one compatible with the new interpolant~\cite{kim2025flowits}:
\[
\bar{u}_s(\bar{x}_s) = \frac{\dot{c}_s}{c_s} \bar{x}_s + c_s \dot{t}_s u_{t_s}(\bar{x}_s / c_s),
\]
where \(c_s = \bar{\sigma}_s / \sigma_{t_s}\) and \(t_s = \rho^{-1}(\bar{\rho}(s))\) is defined via signal-to-noise ratio schedules \(\rho(t) = \alpha_t/\sigma_t\), \(\bar{\rho}(s) = \bar{\alpha}_s/\bar{\sigma}_s\). By converting the flow path to a diffusion path at inference, they increase the diversity of generated samples, which they claim to be the main hindrance to inference-time compute scaling for FM. They couple this novel approach with particle sampling approaches for path selection. This approach loses key benefits of flow matching: the linear interpolant and fewer needed sampling steps at inference due to straighter trajectories. Meanwhile, \textbf{our method preserves the original FM linear interpolant, being the first to demonstrate inference-time compute scaling for the original flow matching formulation}. Further, both inference-time scaling methods here demonstrate improvements only for image generation, while \textbf{we generalize flow matching inference-time scaling to biological problems like protein design and folding.}

\subsection{Noise Injection while Preserving the Continuity Equation}
Existing work aims to inject stochasticity during inference while trying to minimize non-conservative perturbations of the probability density $p_t$ at any time $t$. A widely adopted approach, referred to as EDM (Elucidating the design space of Diffusion Models) introduced by ~\cite{karras2022elucidatingdesignspacediffusionbased}, is commonly used in diffusion models. It introduces noise (attempting to) preserve the marginal densities \(p_t(x)\) \emph{in expectation}. The sampler follows an SDE of the form:
\[
d x_t = u_t(x_t) - \beta(t)\,\sigma^2(t)\,\nabla \log p_t(x_t)\,dt + \sqrt{2\beta(t)}\,\sigma(t)\,dW_t,
\]
Where \(\beta(t)\) is a defined schedule governing the amount of stochasticity injected per step. 

\subsection{Diversity-Promoting Sampling without Retraining Flow Models}
Increasing sample diversity at inference time without retraining can be achieved by coupling concurrently generated particles to avoid redundant generations while retaining fidelity. \textbf{Particle Guidance}~\cite{corso2023particleguidancenoniiddiverse} augments diffusion model sampling dynamics with a joint potential over a cohort of particles, where a fixed similarity kernel (e.g., Euclidean or RBF) induces repulsion that spreads the set across modes while the underlying score dynamics preserve quality. The approach adds modest overhead and is easily layered onto existing samplers. \textbf{DiverseFlows}~\cite{morshed2025diverseflowsampleefficientdiversemode} casts diversity as set coverage via determinantal point processes (DPPs), yielding volume-based gradients that encourage generated sets to span larger regions of the target space. Although training-free, faithfully realizing DPP guidance in high-dimensional image generation requires additional quality-correction terms and nontrivial inter-particle coupling, complicating deployment under strict compute budgets.

\section{Inference Time Scaling for Flow Matching along the Linear Interpolant}

Our investigation of inference-time scaling for FM operates along two independent axes: \textbf{noising schedules} (Sections 4.1-4.2) that introduce stochasticity during sampling, and \textbf{inference algorithms} (Section 4.3) that leverage this stochasticity for search algorithms guided by the verifier function \( r(\cdot) \).

\subsection{Diversity Maximizing Noise Schedule}
\label{sec:custom-rand-ode}
We introduce a randomized-ODE formulation that maximizes the diversity of generated trajectories at inference without reducing the quality of generated samples. While ~\cite{kim2025flowits} convert the original linear flow interpolant to a Variance-Preserving one, we opt to maximize diversity while maintaining the linear interpolant, adhering to the flow matching formulation. We couple time-decaying noise (as we can inject more noise in earlier steps, where the sample is very noisy, than in the later steps where it is closer to a complete sample) with a small particle-guidance repulsion from ~\cite{corso2023particleguidancenoniiddiverse}. Furthermore, we experiment with a score-orthogonal projection of the noise; at inference we use the same procedure as ~\cite{kim2025flowits} to analytically transform the learned velocity \( u_t \) into the score \( \nabla \log p_t(x_t) \). 
\[
dx_t \,=\, \big(u_t(x_t) + \lambda\,w_t(x_t)\big)\,dt,\qquad
w_t(x) \,=\, \alpha(t)\,\Pi_{\perp s_t}\Big[\,\varepsilon\; +\; \eta\,g_t(x)\,\Big],
\]
where $\varepsilon\!\sim\!\mathcal{N}(0,I)$, $\Pi_{\perp s_t}$ projects orthogonally to the learned score direction $s_t = \nabla_x \log p_t(x)$ (see Appendix~\ref{app:score-orthogonal-analysis} for details), and $\alpha(t)$ decays linearly from $1.0$ at $t{=}0$ to $0.7$ at $t{=}1$, scaling the injected noise term. The guidance field $g_t$ uses a fixed kernel potential over the batch (where particles are only part of the same batch when they start from the same initial noise). We set the guidance magnitude $\eta = 0.02$ (2\% of the random Gaussian component), as we find that large particle guidance terms push trajectories too far from the trained paths and degrade sample quality. We set $\eta = 0$ for all protein-design experiments as we find quality degradation to be significant. We opt not to use the DiverseFlows methodology ~\cite{morshed2025diverseflowsampleefficientdiversemode} even though it is specifically designed for FM, as it requires corrector terms to maintain quality, overcomplicating the overall design of our noising schema.

We denote our noising scheme diversity maximizing flow matching ODE (DMFM-ODE).

\subsection{Quality–Diversity Trade-off: Pareto Analysis}
\label{sec:noise_study}

We evaluate how varying the level of stochasticity affects the quality-diversity pareto frontier across our noise schedule (DMFM-ODE) against multiple ablations. We use the pretrained \textsc{SiT-XL/2} model on ImageNet 256×256 with 1,024 samples per configuration. We evaluate on Fréchet Inception Distance (FID) ~\cite{heusel2018fid} and Inception Score (IS) against Inception-v3~\cite{szegedy2015rethinkinginceptionarchitecturecomputer} feature diversity (measured as average batch distance, where each element in the batch begins at the same initial noise). We vary the noise magnitude to create multiple points for each method on the pareto frontier. The methods are:
\begin{itemize}
  \item \textbf{DMFM-ODE (ours)}: Composite randomized ODE (Section~\ref{sec:custom-rand-ode}) with time-decayed noise and weak particle guidance.
  \item \textbf{ODE-scoreorth}: score-orthogonal perturbations that minimize probability density shifts, where \( \lambda \) controls the noise magnitude.
  \[
  dx_t = \left(u_t(x) + \lambda\,\Pi_{\perp s_t}[\varepsilon]\right) dt,
  \]
  \item \textbf{SDE}: Samples are generated from a simple Euler-Maruyama SDE.   where \( dW_t \sim \mathcal{N}(0, dt) \), and \(\sigma\) scales the noise magnitude.
  \[
  dx_t = u_t(x)\,dt + \sigma\,dW_t,
  \]
  \item \textbf{EDM-SDE}: A stochastic method that adjusts both drift and diffusion to preserve \(p_t(x)\) in expectation. where \(\beta(t)\) is a user-defined schedule. See Related Work for details (Section~\ref{sec:related}):
  \[
  dx_t = u_t(x)\,dt - \beta(t)\,\sigma^2(t)\,\nabla \log p_t(x)\,dt + \sqrt{2\beta(t)}\,\sigma(t)\,dW_t,
  \]
  \item \textbf{Score-SDE}: SDE which denoises via the score (computed analytically from the velocity at inference) as defined in ~\cite{kim2025flowits} (see Related Work section ~\ref{sec:related-fmscaling} for full definition).
\end{itemize}

\begin{figure}[H]
  \centering
  \includegraphics[width=0.6\textwidth]{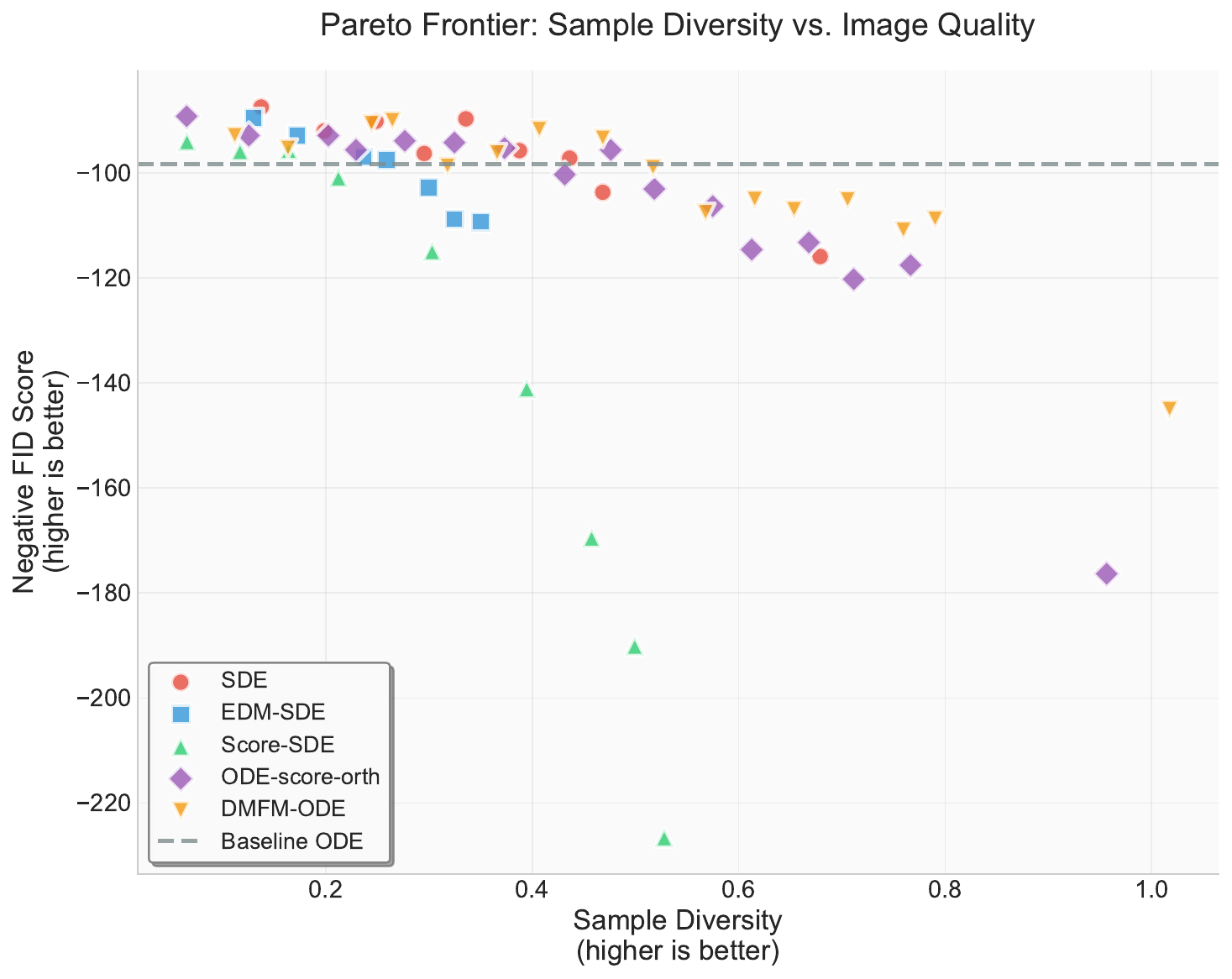}
  \caption{Pareto frontier: sample diversity vs. negative FID (higher is better on both axes). The composite DMFM-ODE expands the frontier relative to ablations and SDE baselines.}
  \label{fig:pareto-fid}
\end{figure}

Figure ~\ref{fig:pareto-fid} demonstrates that DMFM-ODE pushes the pareto front outward at higher noise levels, enabling more stochasticity (and thus more diversity) before quality degrades substantially. We can therefore select a higher noise magnitude for DMFM-ODE than baselines such as the Euler-Maruyama SDE. An identical plot for IS is in Appendix~\ref{app:noise-diversity-results}, with additional noise curves and experimental details.

\subsection{Inference Algorithms}

We introduce three inference-time compute scaling algorithms that leverage stochasticity to search for high-quality samples under the verifier $r(\cdot)$. Detailed algorithmic descriptions and pseudocode for all methods are provided in Appendix~\ref{sec:algorithms}.

\textbf{Random Search (Best-of-N)} generates $N$ independent samples via deterministic ODE Euler sampling from random initial conditions and returns the top-$K$ samples as determined by $r(\cdot)$. We adopt the "Random Search" (RS) terminology of ~\cite{ma2025diffits}. 

\textbf{Noise Search (NS, ours)} stochastically samples $N$ candidates from identical initial/intermediate conditions at time $t$ (the trajectories are saved) for round $i$. The top-$K$ samples, determined by the verifier $r(\cdot)$, are used in the next round $i+1$, where we take the trajectory of the chosen samples in round $i$ at $startTime_{i+1}$ as the starting condition for beginning sampling at $t=startTime_{i+1}$. In this way we refine the final trajectory iteratively towards the optimal and verify exact final samples with $r(\cdot)$, while maintaining tractability. In practice $N$ is our compute scaling factor, we set $k=1$ and $startTime=[0.0, 0.2, 0.4, 0.6, 0.75, 0.8, 0.85, 0.9, 0.95]$. Complete implementation details are provided in Appendix~\ref{app:implementation-details}.

\textbf{RS + Noise Search (RS+NS, ours)} first performs random search over initial noise conditions, then uses the best candidate initial noises as the initial noises at round $i=0$, where $startTime_{i=0}=0$. This highlights an advantage of our noise search algorithm, namely that it optimizes the trajectory on $t \in [0,1]$, but is invariant to the initial noise (or more generally for flow matching, the distribution at $t=0$). We can therefore optimize the sample by selecting an optimal initial noise, then independently optimize the sample further through the selection of the trajectory, conditional on the initial noises selected.

\section{Experiments}
We validate our methods on ImageNet (256x256) generation and unconditional protein structure generation, using the pretrained \textsc{SiT-XL/2} and \textsc{FoldFlow2} models, respectively. We evaluate each method with inference scaling factors of $N=$ 1 (base model ODE Euler sampling, no inference scaling), 2, 4, 8, and assume access to the verifier $r(\cdot)$ is fast and unlimited. We evaluate DMFM noise alongside the Noise Search \textbf{(NS–DMFM-ODE, ours)} and two-stage \textbf{(RS+NS–DMFM-ODE, ours)} inference algorithms, with Random Search \textbf{(RS)} performing as a competitive baseline. We include the Euler-Maruyama noise schedule with Noise Search \textbf{(NS–SDE)} as an ablation, allowing us to validate how increasing sample diversity affects scaling performance. Complete experimental details and hyperparameters are provided in Appendix~\ref{app:implementation-details}.

\subsection{ImageNet 256x256}
We evaluate on ImageNet 256×256 using the pretrained \textsc{SiT-XL/2} flow-matching model. All evaluations generate 1,024 samples per method, per scaling factor. Following the experimental design of ~\cite{ma2025diffits}, we use both Inception Score (IS) ~\cite{salimans2016inception} and DinoV2 classification accuracy (CA) ~\cite{oquab2024dinov2learningrobustvisual} as verifier functions $r(\cdot)$. Performance is measured using FID ~\cite{heusel2018fid}, IS, and DinoV2 CA (top-1).

\begin{figure}[H]
  \centering
  \includegraphics[width=0.9\textwidth]{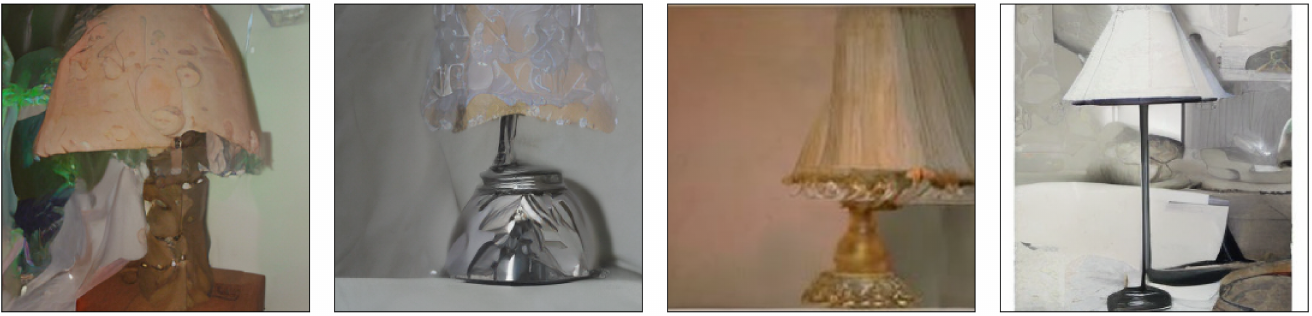}
  \caption{Sample quality improvement with inference-time compute scaling using our RS+NS–DMFM-ODE method on ImageNet 256×256 with IS verifier. From left to right: 1×, 2×, 4×, 8× compute budget.}
  \label{fig:inception-two-stage-samples}
\end{figure}

\begin{figure}[H]
  \centering
  \begin{minipage}{0.48\textwidth}
    \centering
    \includegraphics[width=\textwidth]{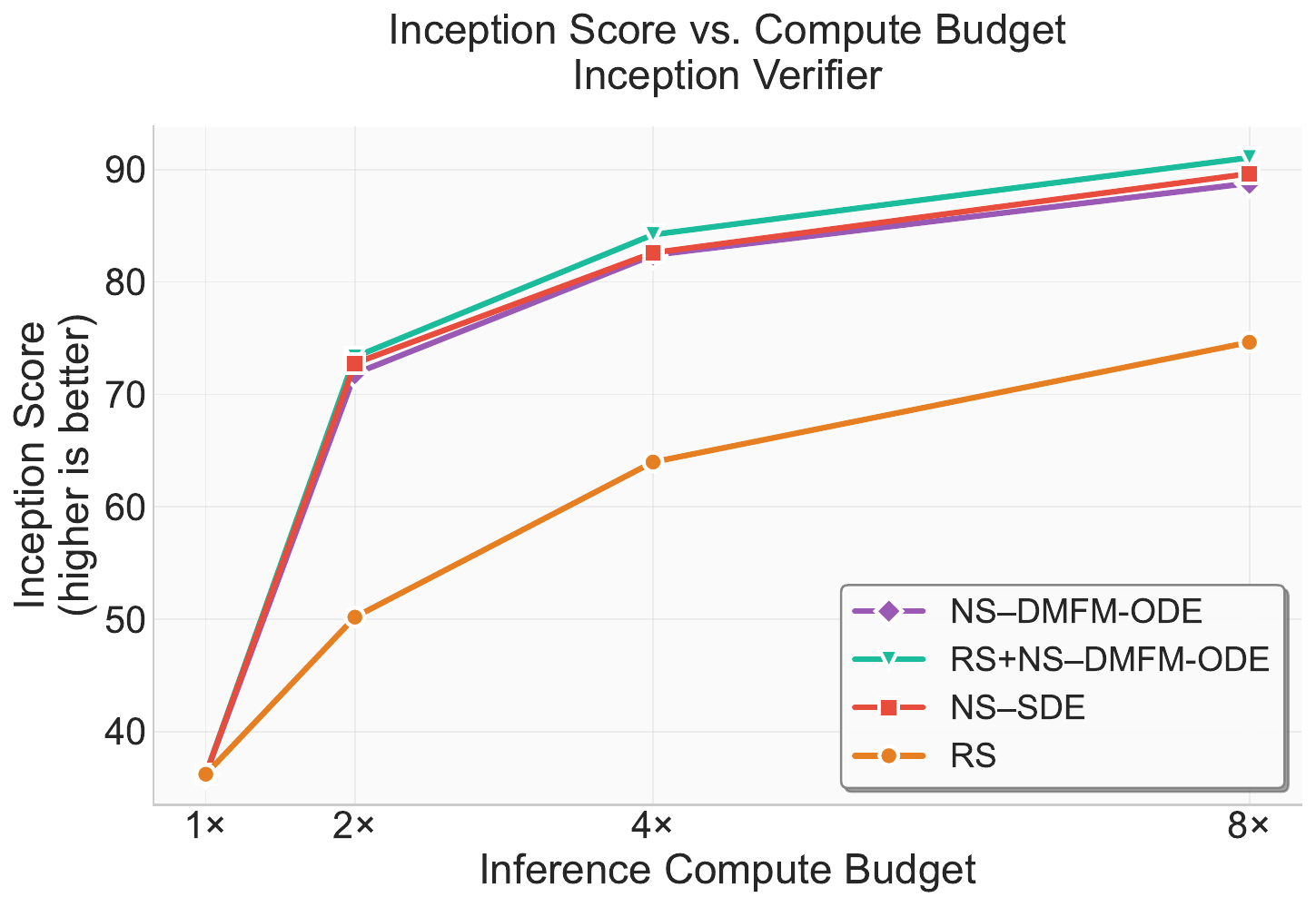}
  \end{minipage}
  \hfill
  \begin{minipage}{0.48\textwidth}
    \centering
    \includegraphics[width=\textwidth]{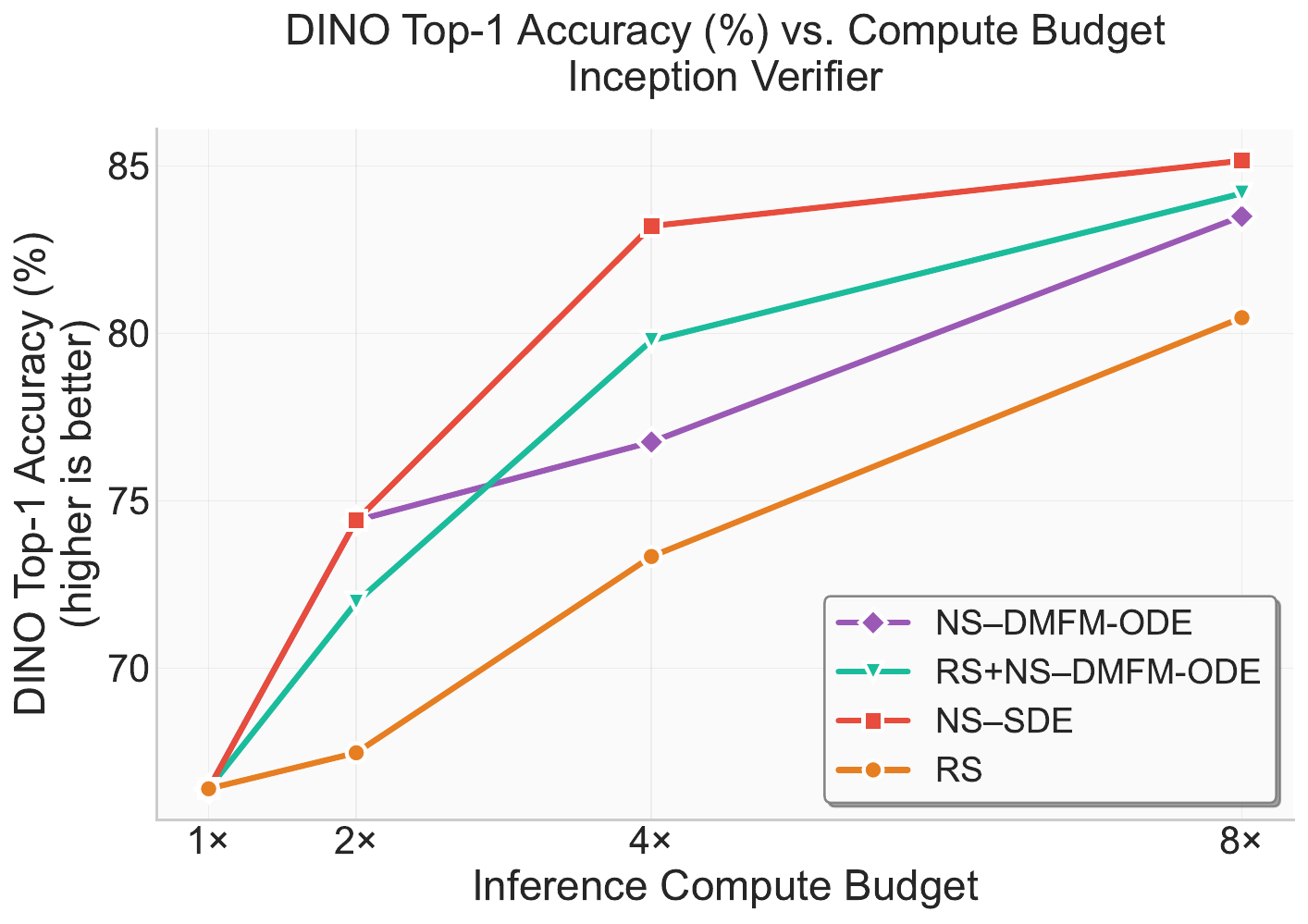}
  \end{minipage}
  \caption{Inference-time scaling results using IS as verifier. Left: Inception Score vs. Scaling Factor. Right: DINO Top-1 accuracy vs. Scaling Factor.}
  \label{fig:inception-scaling}
\end{figure}

Results for inference time scaling with the Inception Score as the verifier function are visible in Figure ~\ref{fig:inception-scaling}.\textbf{ All three noise search methods outperform Random Search, with the two stage RS + DMFM noise search (RS+NS–DMFM-ODE) having the highest performance} (but only marginally) when looking at Inception Score itself. The SDE noise search and DMFM (ours) perform equally. Interestingly the SDE noise search shows the largest improvements in DINO classification accuracy when IS is used as the verifier. We also present performance of IS guided scaling on the DINO top-5 CA and FID metrics in Appendix ~\ref{app:complete-scaling-results}, though less meaningful (scaling on the IS does not produce meaningful reductions in the FID).

\begin{figure}[H]
  \centering
  \begin{minipage}{0.32\textwidth}
    \centering
    \includegraphics[width=\textwidth]{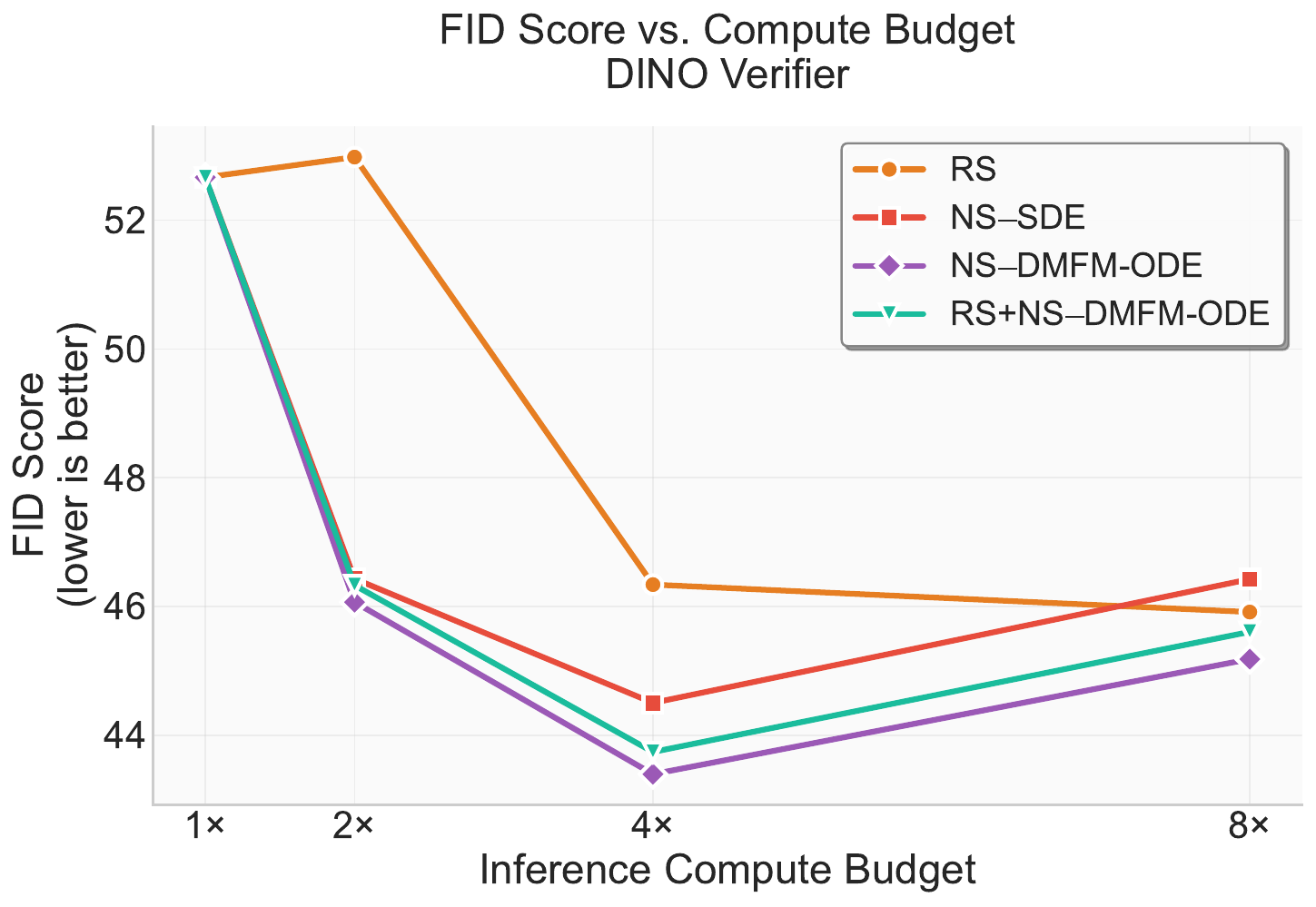}
  \end{minipage}
  \hfill
  \begin{minipage}{0.32\textwidth}
    \centering
    \includegraphics[width=\textwidth]{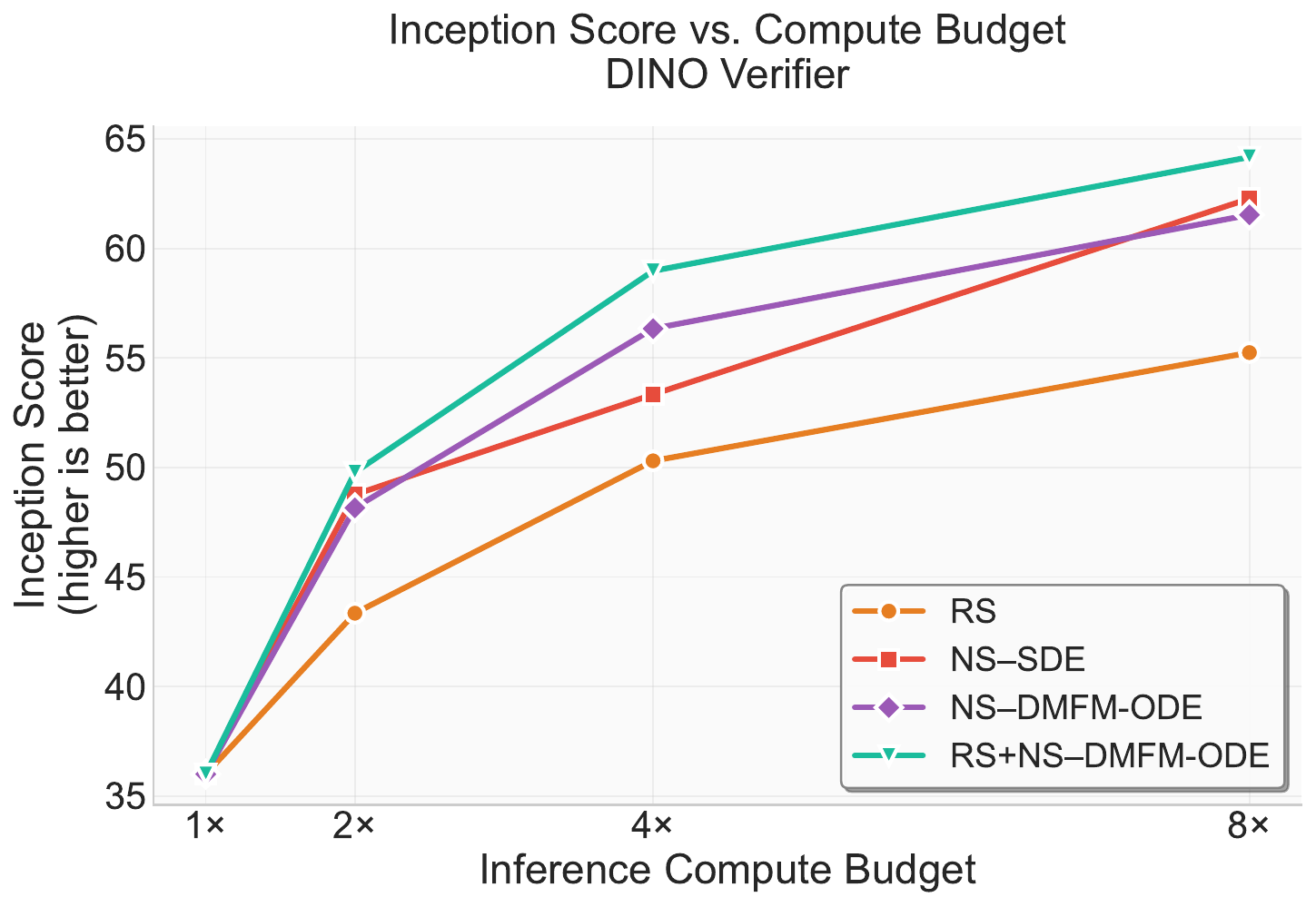}
  \end{minipage}
  \hfill
  \begin{minipage}{0.32\textwidth}
    \centering
    \includegraphics[width=\textwidth]{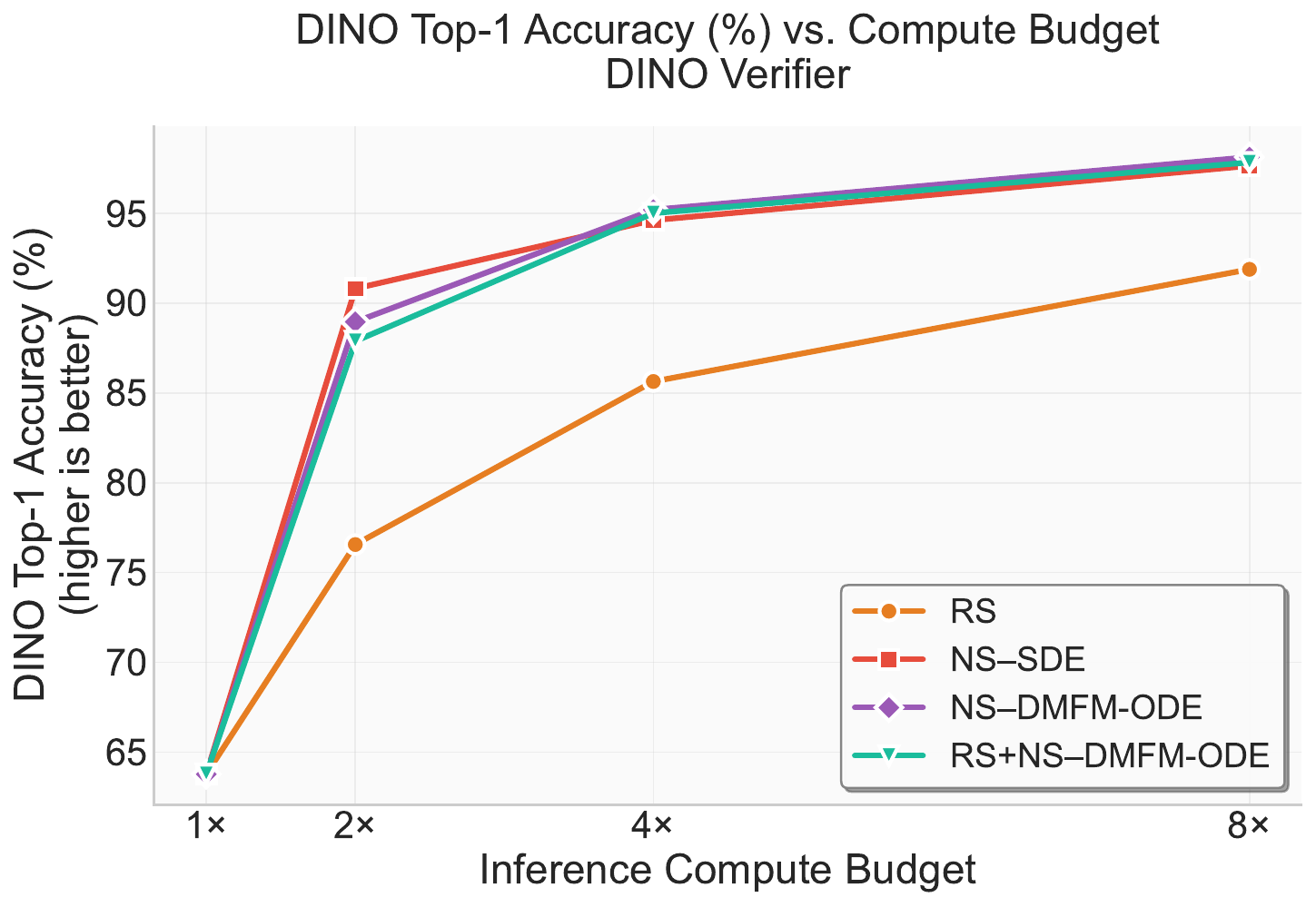}
  \end{minipage}
  \caption{Inference-time scaling results using DinoV2 as verifier. Left: FID vs. Scaling Factor. Center: Inception Score vs. Scaling Factor. Right: DINO Top-1 accuracy vs. Scaling Factor.}
  \label{fig:dino-scaling}
\end{figure}

Again, when scaling with DINO CA (top-1) as a verifier in Figure \ref{fig:dino-scaling}, we see that all methods outperform Random Search. As an oracle verifier (i.e increasing DINO top-1 accuracy with DINO as the verifier function), the Noise Search methods approach 100 percent classification accuracy. Scaling on the DINO CA also translates to reductions in the FID, with the two stage method (RS + NS-DMFM-ODE) showing clearer performance gains in terms of Inception Score. Remaining metrics under DINO-guided scaling are provided in Appendix~\ref{app:complete-scaling-results}. Additional visual examples of Random Search and RS+NS–DMFM-ODE sampling across different compute budgets are shown in Appendix~\ref{app:additional-visual-examples}.

\subsection{FoldFlow: Protein Design}

We use the pretrained FoldFlow2 model~\cite{foldflow2} for the task of unconditional protein structure generation. To our knowledge, \textbf{this is the first demonstration of inference-time scaling for flow matching applied to protein design}. We generate 64 protein samples of length 100 residues per configuration. Following the FoldFlow evaluation framework, we use self-consistency template modeling score (scTM-score) as the verifier function for sample selection, which optimizes for protein \textit{designability}. self-consistency refers to a folding-refolding cycle: FoldFlow generates a backbone structure, ProteinMPNN~\cite{proteinmpnn} predicts 8 amino acid sequences for that structure, ESMFold~\cite{esmfold} refolds these sequences into 3D structures, and structural alignment metrics (TM-score and Root Mean Square Deviation, RMSD) compare the refolded structures to the original. A high self-consistency score means that structure likely has a realistic, natural, sequence-structure relationship. We evaluate performance using scTM-score, self-consistency RMSD (scRMSD), and hard thresholds measuring the percentage of samples with scRMSD below 2.0Å (the \textit{designability} threshold of ~\cite{foldflow2}). Detailed FoldFlow experimental parameters are provided in Appendix~\ref{app:foldflow-details}.

\newpage

\begin{figure}[H]
  \centering
  \begin{minipage}{0.32\textwidth}
    \centering
    \includegraphics[width=\textwidth]{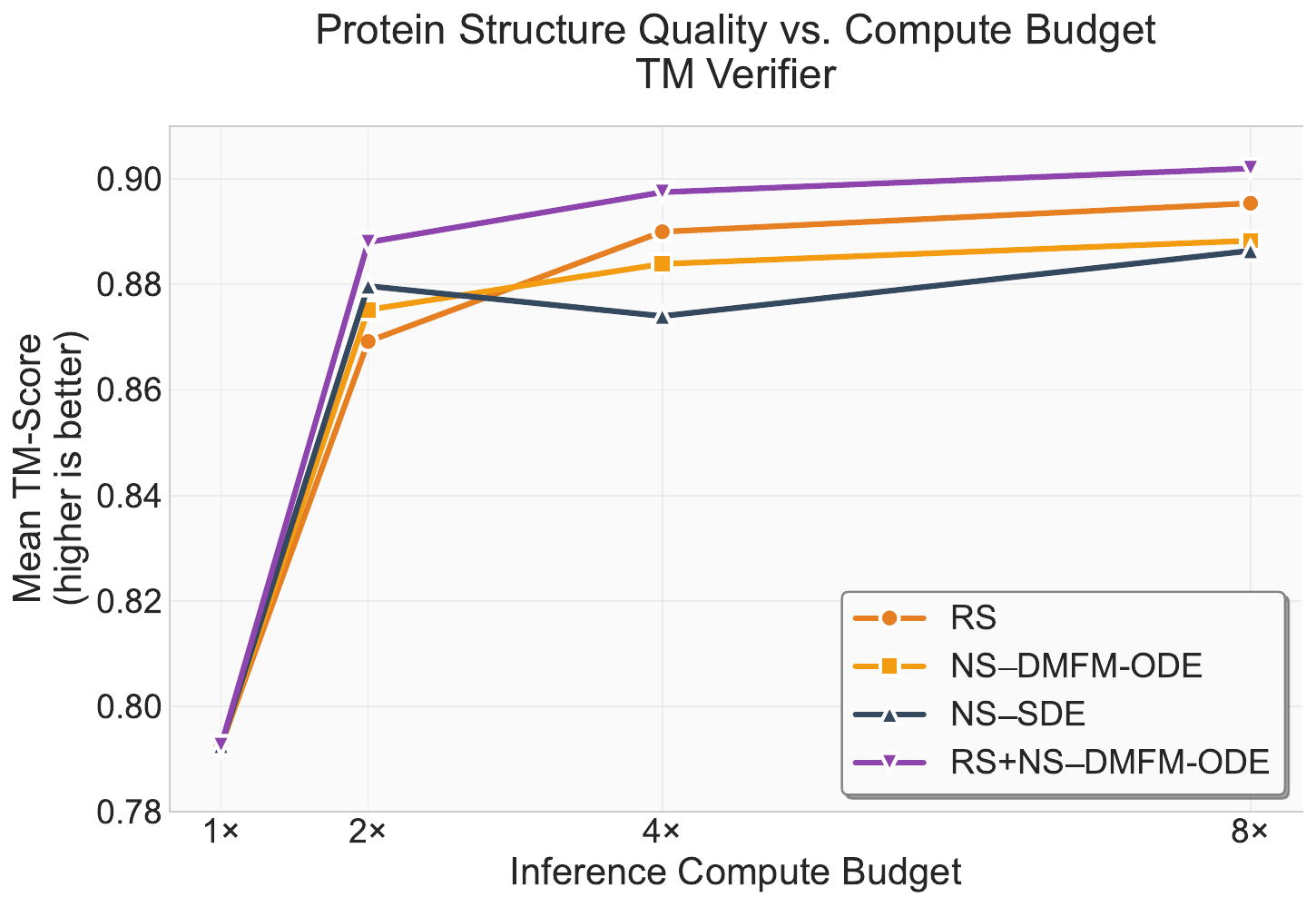}
  \end{minipage}
  \hfill
  \begin{minipage}{0.32\textwidth}
    \centering
    \includegraphics[width=\textwidth]{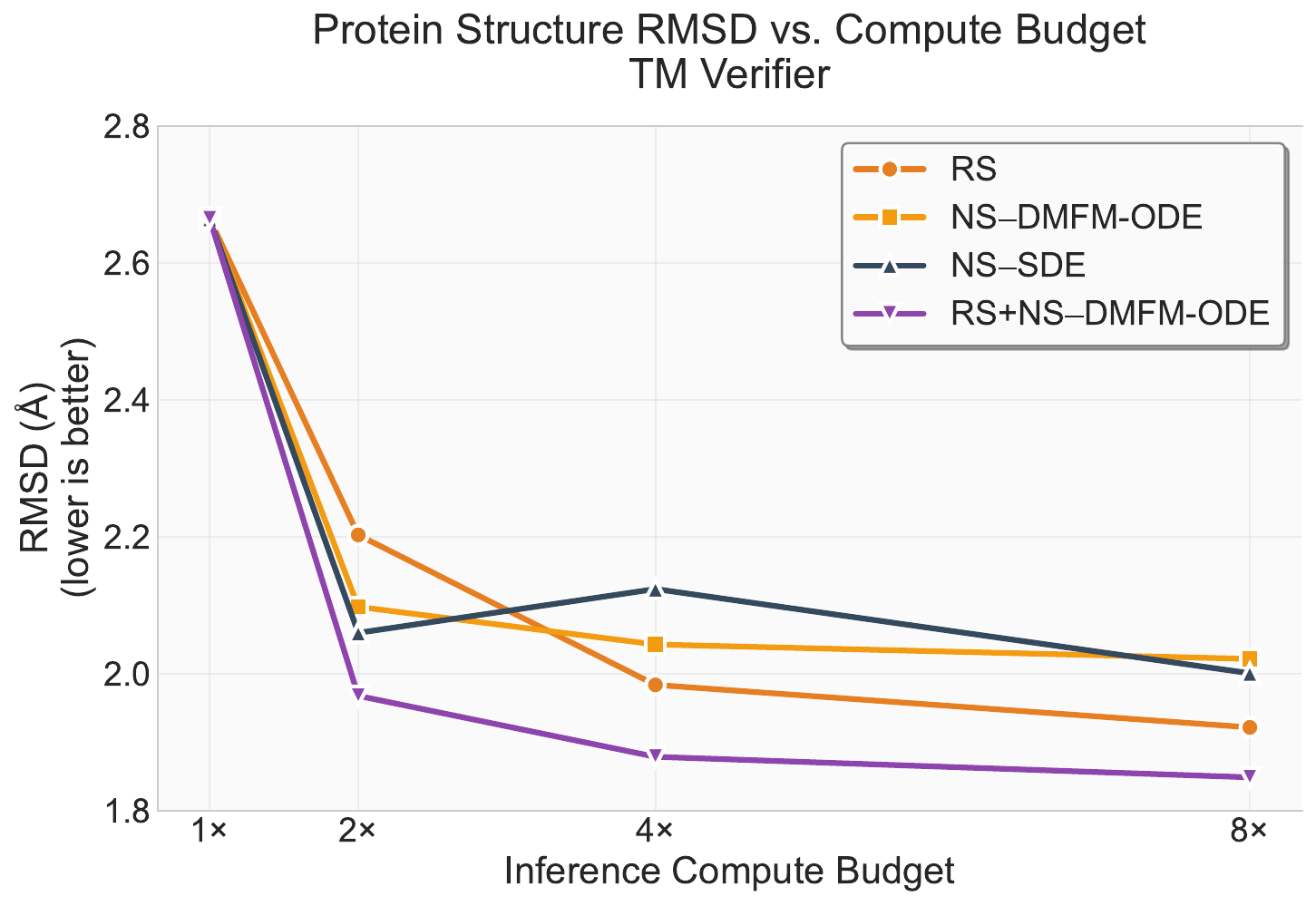}
  \end{minipage}
  \hfill
  \begin{minipage}{0.32\textwidth}
    \centering
    \includegraphics[width=\textwidth]{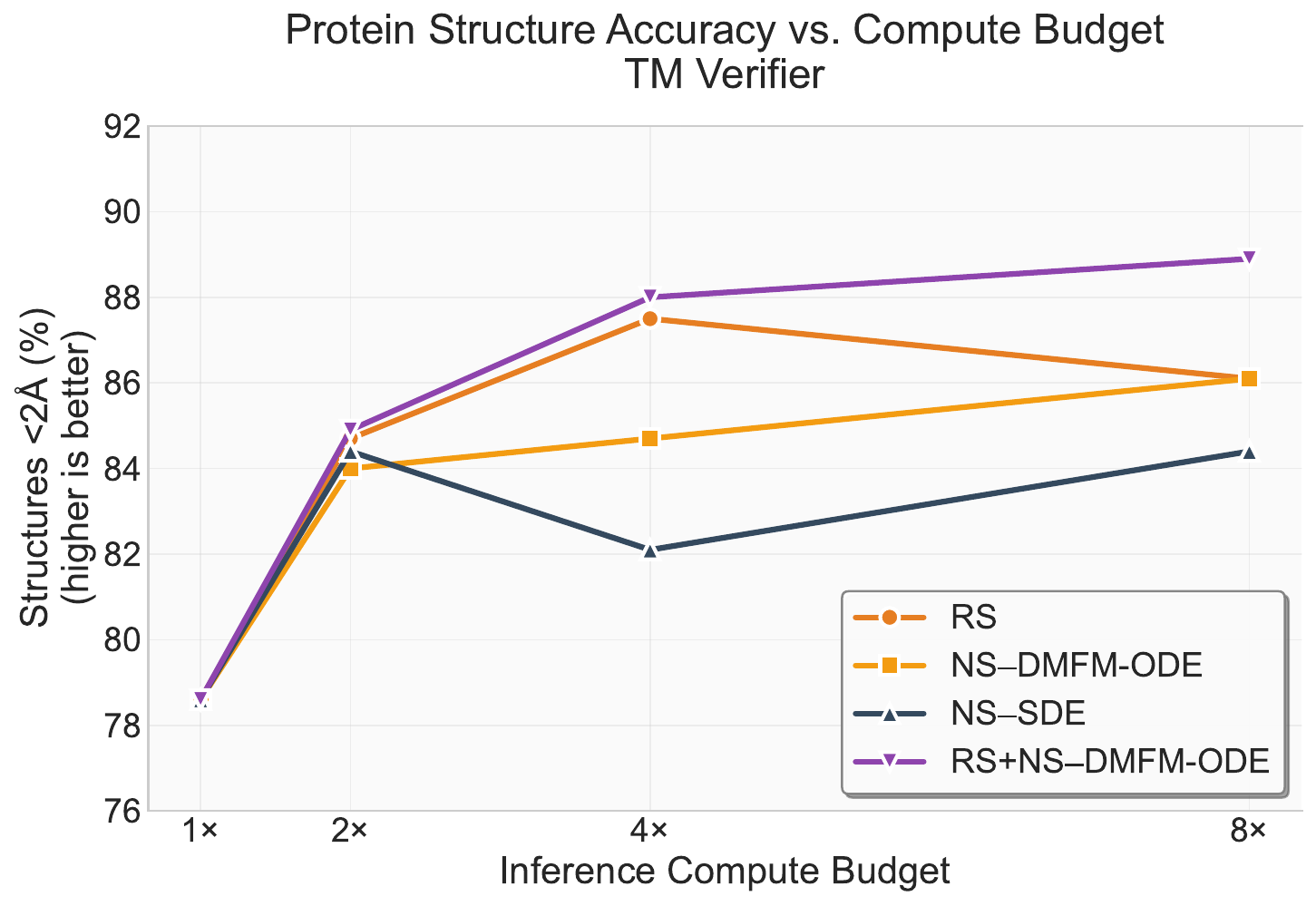}
  \end{minipage}
  \caption{Protein generation results using TM verifier function. Left: TM-score vs. compute budget. Center: RMSD vs. compute budget (lower is better). Right: Percentage of structures with RMSD $<$ 2Å vs. compute budget.}
  \label{fig:protein-scaling}
\end{figure}

The results in Figure \ref{fig:protein-scaling} demonstrate substantial improvements across all metrics as the compute budget increases. The two-stage \textbf{RS+NS–DMFM-ODE method achieves the highest performance, being the only method with an average TM-score above 0.9 at 8x compute}. Here we can clearly justify our two stage method as a substantial improvement over Random Search and Noise search being used independently. SDE Noise Search (NS-SDE) has the worst performance, potentially implying it suffers from a lack of diversity. The higher levels of noise that DMFM-ODE allows us to inject at inference (without reducing quality, as we demonstrated in Section 4) add more diversity and thus appear more likely to lead to larger gains when additional compute is applied. We include a similar analysis for a \textbf{geometric verifier function} in Appendix~\ref{app:geometric-scoring}, using fast and easily computable geometric properties (Ca distance, number of clashes, etc.). However the results appear uncorrelated, implying that optimizing for the geometric verifier function does not optimize protein \textit{designability}.

\subsection{Interpretation}
\textbf{Verifier-guided inference-time scaling translates well to flow matching via the Noise Search algorithm.} Results demonstrate monotonically increasing performance as the number of search branches is increased. This gain tends to plateau as we scale to 8x compute. On the image generation task, Noise Search substantially outperforms the Random Search baseline, with competitive results on protein generation

\textbf{Verifier-guided inference-time scaling for flow matching translates to biological tasks such as protein design.} We show protein \textit{designability} can be improved via our Noise Search method, with our two-stage algorithm achieving better performance than the Random Search baseline. 

\textbf{Taking advantage of Flow Matching's prior distribution invariance is critical to further boost performance through a multi-stage algorithm.} We see that the two stage \textbf{RS+NS-DMFM-ODE} method achieves the highest scores on the image generation task across almost all metrics, and on all metrics on protein design. This highlights the invariance of our model to initial distributions, a key advantage of flow matching over diffusion, allowing us to first select the initial noises independently of choosing the optimal trajectories.

\textbf{Increasing diversity does not necessarily increase performance in all domains.} We find that at inference, the linear interpolant of flow matching \textit{does} provide sufficient sample diversity to improve performance using the Noise Search algorithm on both the image and protein domains, and in fact outperforms the random search method (which has higher diversity). We find that increasing diversity via the score-orthogonal DMFM noise schedule does not necessarily improve performance versus the SDE baseline in the image domain, but potentially does in the protein design task. We speculate that finding very different samples is important for wide exploration, but exploiting well performing samples is more lucrative, while iteratively doing both (exploration then exploitation) has the highest performance gain.

\section{Conclusion}
We demonstrate the first application of inference-time compute scaling to flow matching, maintaining the linear interpolant at inference, with our noise search algorithm. Our approach outperforms baselines on two domains which flow matching excels, image generation, and for the first time, protein design, showing substantial gains as compute increases. We introduce a noise schedule that maximizes the diversity-quality pareto frontier relative to a standard SDE. However we find that increasing diversity may not necessarily benefit search-scaling algorithms, with evidence that it is specific to the domain. We additionally introduce a two-stage algorithm that achieves state of the art results by first searching over initial noises with random search, then refining trajectories with noise search. This joint algorithm is enabled by the fact that noise search method is agnostic to the initial noise condition (or more generally to the initial sample at $t=0$), but this property may also allow us to apply inference-time scaling to problems where the distribution $p_0$ is not a simple gaussian, but is itself a dataset such as in the case of cell trajectories, which is also of scientific interest. Validating our method on this problem is one area we leave to future work.

\textbf{Future work and limitations.} Aside from the above, we believe that our method, as well as other inference-time compute scaling methods for stochastic interpolants, are fundamentally limited by the pretrained models they use. As these models are not explicitly trained for inference-time scaling, improvements are limited and the rate of improvement decreases as compute scales. While flow matching is designed for efficient sampling, there is enormous gain from slower higher quality sampling for scientific discovery, such as protein design for therapeutics. We believe that future work should experiment with models trained to "think" longer, equivalent to chain-of-thought in LLMs ~\cite{openai2024o1}, or learn to incorporate prior information during sampling. This, combined with the methods we have introduced, may enable flow matching methods that require more compute but achieve results far beyond what is currently possible.

\section{Acknowledgments}

The authors would like to acknowledge funding from McGill, MILA, CIFAR and NSERC. The research was enabled in part by computational resources provided by the Digital Research Alliance of Canada (\url{https://alliancecan.ca}), Mila (\url{https://mila.quebec}), Nvidia, and by free compute credits provided by Lambda Compute (\url{https://lambda.ai/}). We'd like to thank Lucas Nelson, Lazar Atanackovic, and Alex Tong, for their valuable feedback and discussions.

\section*{Reproducibility Statement}
To ensure reproducibility of our results, we provide comprehensive implementation details and experimental configurations throughout this work. Section~\ref{sec:algorithms} in the appendix contains detailed algorithmic descriptions with pseudocode for all three inference methods (Random Search, Noise Search, and RS+NS). Complete experimental hyperparameters for both ImageNet and FoldFlow experiments are documented in Appendix~\ref{app:implementation-details}, including model configurations, sampling parameters, and verifier settings. The mathematical formulation of our DMFM-ODE noise schedule is provided in Section~\ref{sec:custom-rand-ode}, with score-orthogonal projection analysis detailed in Appendix~\ref{app:score-orthogonal-analysis}. All experiments use publicly available pretrained models (SiT-XL/2 for ImageNet and FoldFlow2 for proteins) with standard evaluation metrics (FID, Inception Score, DINO classification accuracy, TM-score, RMSD). Dataset preprocessing follows established protocols for ImageNet 256×256 and protein structure generation as described in the respective model papers. All code and datasets are publicly available.

\bibliography{main}

\begin{thebibliography}{26}
\providecommand{\natexlab}[1]{#1}
\providecommand{\url}[1]{\texttt{#1}}
\expandafter\ifx\csname urlstyle\endcsname\relax
  \providecommand{\doi}[1]{doi: #1}\else
  \providecommand{\doi}{doi: \begingroup \urlstyle{rm}\Url}\fi

\bibitem[Albergo et~al.(2023)Albergo, Boffi, and Vanden-Eijnden]{albergo2023stochastic_interpolants}
Michael~S. Albergo, Nicholas~M. Boffi, and Eric Vanden-Eijnden.
\newblock Stochastic interpolants: A unifying framework for flows and diffusions, 2023.
\newblock URL \url{https://arxiv.org/abs/2303.08797}.

\bibitem[Brown et~al.(2024)]{brown2024llmmonkeys}
Benjamin Brown et~al.
\newblock Large language monkeys: Scaling inference compute with repeated sampling.
\newblock \emph{arXiv:2407.21787}, 2024.

\bibitem[Cobbe et~al.(2021)]{cobbe2021verifiers}
Karl Cobbe et~al.
\newblock Training verifiers to solve math word problems.
\newblock \emph{arXiv:2110.14168}, 2021.

\bibitem[Corso et~al.(2023)Corso, Xu, de~Bortoli, Barzilay, and Jaakkola]{corso2023particleguidancenoniiddiverse}
Gabriele Corso, Yilun Xu, Valentin de~Bortoli, Regina Barzilay, and Tommi Jaakkola.
\newblock Particle guidance: non-i.i.d. diverse sampling with diffusion models, 2023.
\newblock URL \url{https://arxiv.org/abs/2310.13102}.

\bibitem[Dauparas et~al.(2022)Dauparas, Anishchenko, Bennett, Bai, Ragotte, Milles, Wicky, Courbet, de~Haas, Bethel, Leung, Huddy, Pellock, Tischer, Chan, Koepnick, Nguyen, Kang, Sankaran, Bera, King, and Baker]{proteinmpnn}
J.~Dauparas, I.~Anishchenko, N.~Bennett, H.~Bai, R.~J. Ragotte, L.~F. Milles, B.~I.~M. Wicky, A.~Courbet, R.~J. de~Haas, N.~Bethel, P.~J.~Y. Leung, T.~F. Huddy, S.~Pellock, D.~Tischer, F.~Chan, B.~Koepnick, H.~Nguyen, A.~Kang, B.~Sankaran, A.~K. Bera, N.~P. King, and D.~Baker.
\newblock Robust deep learning–based protein sequence design using proteinmpnn.
\newblock \emph{Science}, 378\penalty0 (6615):\penalty0 49--56, 2022.
\newblock \doi{10.1126/science.add2187}.
\newblock URL \url{https://www.science.org/doi/abs/10.1126/science.add2187}.

\bibitem[DeepSeek-AI et~al.(2025)DeepSeek-AI, Guo, Yang, Zhang, Song, Zhang, Xu, Zhu, Ma, Wang, Bi, Zhang, Yu, Wu, Wu, Gou, Shao, Li, Gao, Liu, Xue, Wang, Wu, Feng, Lu, Zhao, Deng, Zhang, Ruan, Dai, Chen, Ji, Li, Lin, Dai, Luo, Hao, Chen, Li, Zhang, Bao, Xu, Wang, Ding, Xin, Gao, Qu, Li, Guo, Li, Wang, Chen, Yuan, Qiu, Li, Cai, Ni, Liang, Chen, Dong, Hu, Gao, Guan, Huang, Yu, Wang, Zhang, Zhao, Wang, Zhang, Xu, Xia, Zhang, Zhang, Tang, Li, Wang, Li, Tian, Huang, Zhang, Wang, Chen, Du, Ge, Zhang, Pan, Wang, Chen, Jin, Chen, Lu, Zhou, Chen, Ye, Wang, Yu, Zhou, Pan, Li, Zhou, Wu, Ye, Yun, Pei, Sun, Wang, Zeng, Zhao, Liu, Liang, Gao, Yu, Zhang, Xiao, An, Liu, Wang, Chen, Nie, Cheng, Liu, Xie, Liu, Yang, Li, Su, Lin, Li, Jin, Shen, Chen, Sun, Wang, Song, Zhou, Wang, Shan, Li, Wang, Wei, Zhang, Xu, Li, Zhao, Sun, Wang, Yu, Zhang, Shi, Xiong, He, Piao, Wang, Tan, Ma, Liu, Guo, Ou, Wang, Gong, Zou, He, Xiong, Luo, You, Liu, Zhou, Zhu, Xu, Huang, Li, Zheng, Zhu, Ma, Tang, Zha, Yan, Ren, Ren, Sha, Fu, Xu, Xie, Zhang,
  Hao, Ma, Yan, Wu, Gu, Zhu, Liu, Li, Xie, Song, Pan, Huang, Xu, Zhang, and Zhang]{deepseek2025r1}
DeepSeek-AI, Daya Guo, Dejian Yang, Haowei Zhang, Junxiao Song, Ruoyu Zhang, Runxin Xu, Qihao Zhu, Shirong Ma, Peiyi Wang, Xiao Bi, Xiaokang Zhang, Xingkai Yu, Yu~Wu, Z.~F. Wu, Zhibin Gou, Zhihong Shao, Zhuoshu Li, Ziyi Gao, Aixin Liu, Bing Xue, Bingxuan Wang, Bochao Wu, Bei Feng, Chengda Lu, Chenggang Zhao, Chengqi Deng, Chenyu Zhang, Chong Ruan, Damai Dai, Deli Chen, Dongjie Ji, Erhang Li, Fangyun Lin, Fucong Dai, Fuli Luo, Guangbo Hao, Guanting Chen, Guowei Li, H.~Zhang, Han Bao, Hanwei Xu, Haocheng Wang, Honghui Ding, Huajian Xin, Huazuo Gao, Hui Qu, Hui Li, Jianzhong Guo, Jiashi Li, Jiawei Wang, Jingchang Chen, Jingyang Yuan, Junjie Qiu, Junlong Li, J.~L. Cai, Jiaqi Ni, Jian Liang, Jin Chen, Kai Dong, Kai Hu, Kaige Gao, Kang Guan, Kexin Huang, Kuai Yu, Lean Wang, Lecong Zhang, Liang Zhao, Litong Wang, Liyue Zhang, Lei Xu, Leyi Xia, Mingchuan Zhang, Minghua Zhang, Minghui Tang, Meng Li, Miaojun Wang, Mingming Li, Ning Tian, Panpan Huang, Peng Zhang, Qiancheng Wang, Qinyu Chen, Qiushi Du, Ruiqi Ge, Ruisong
  Zhang, Ruizhe Pan, Runji Wang, R.~J. Chen, R.~L. Jin, Ruyi Chen, Shanghao Lu, Shangyan Zhou, Shanhuang Chen, Shengfeng Ye, Shiyu Wang, Shuiping Yu, Shunfeng Zhou, Shuting Pan, S.~S. Li, Shuang Zhou, Shaoqing Wu, Shengfeng Ye, Tao Yun, Tian Pei, Tianyu Sun, T.~Wang, Wangding Zeng, Wanjia Zhao, Wen Liu, Wenfeng Liang, Wenjun Gao, Wenqin Yu, Wentao Zhang, W.~L. Xiao, Wei An, Xiaodong Liu, Xiaohan Wang, Xiaokang Chen, Xiaotao Nie, Xin Cheng, Xin Liu, Xin Xie, Xingchao Liu, Xinyu Yang, Xinyuan Li, Xuecheng Su, Xuheng Lin, X.~Q. Li, Xiangyue Jin, Xiaojin Shen, Xiaosha Chen, Xiaowen Sun, Xiaoxiang Wang, Xinnan Song, Xinyi Zhou, Xianzu Wang, Xinxia Shan, Y.~K. Li, Y.~Q. Wang, Y.~X. Wei, Yang Zhang, Yanhong Xu, Yao Li, Yao Zhao, Yaofeng Sun, Yaohui Wang, Yi~Yu, Yichao Zhang, Yifan Shi, Yiliang Xiong, Ying He, Yishi Piao, Yisong Wang, Yixuan Tan, Yiyang Ma, Yiyuan Liu, Yongqiang Guo, Yuan Ou, Yuduan Wang, Yue Gong, Yuheng Zou, Yujia He, Yunfan Xiong, Yuxiang Luo, Yuxiang You, Yuxuan Liu, Yuyang Zhou, Y.~X. Zhu,
  Yanhong Xu, Yanping Huang, Yaohui Li, Yi~Zheng, Yuchen Zhu, Yunxian Ma, Ying Tang, Yukun Zha, Yuting Yan, Z.~Z. Ren, Zehui Ren, Zhangli Sha, Zhe Fu, Zhean Xu, Zhenda Xie, Zhengyan Zhang, Zhewen Hao, Zhicheng Ma, Zhigang Yan, Zhiyu Wu, Zihui Gu, Zijia Zhu, Zijun Liu, Zilin Li, Ziwei Xie, Ziyang Song, Zizheng Pan, Zhen Huang, Zhipeng Xu, Zhongyu Zhang, and Zhen Zhang.
\newblock Deepseek-r1: Incentivizing reasoning capability in llms via reinforcement learning, 2025.
\newblock URL \url{https://arxiv.org/abs/2501.12948}.

\bibitem[Dunn \& Koes(2025)Dunn and Koes]{flowmol3}
Ian Dunn and David~R. Koes.
\newblock Flowmol3: Flow matching for 3d de novo small-molecule generation, 2025.
\newblock URL \url{https://arxiv.org/abs/2508.12629}.

\bibitem[Gandhi et~al.(2024)]{gandhi2024sos}
Kanishka Gandhi et~al.
\newblock Stream of search (sos): Learning to search in language.
\newblock \emph{arXiv:2404.03683}, 2024.

\bibitem[Heusel et~al.(2018)Heusel, Ramsauer, Unterthiner, Nessler, and Hochreiter]{heusel2018fid}
Martin Heusel, Hubert Ramsauer, Thomas Unterthiner, Bernhard Nessler, and Sepp Hochreiter.
\newblock Gans trained by a two time-scale update rule converge to a local nash equilibrium, 2018.
\newblock URL \url{https://arxiv.org/abs/1706.08500}.

\bibitem[Huguet et~al.(2024)Huguet, Vuckovic, Fatras, Thibodeau-Laufer, Lemos, Islam, Liu, Rector-Brooks, Akhound-Sadegh, Bronstein, Tong, and Bose]{foldflow2}
Guillaume Huguet, James Vuckovic, Kilian Fatras, Eric Thibodeau-Laufer, Pablo Lemos, Riashat Islam, Cheng-Hao Liu, Jarrid Rector-Brooks, Tara Akhound-Sadegh, Michael Bronstein, Alexander Tong, and Avishek~Joey Bose.
\newblock Sequence-augmented se(3)-flow matching for conditional protein backbone generation, 2024.
\newblock URL \url{https://arxiv.org/abs/2405.20313}.

\bibitem[Karras et~al.(2022)Karras, Aittala, Aila, and Laine]{karras2022elucidatingdesignspacediffusionbased}
Tero Karras, Miika Aittala, Timo Aila, and Samuli Laine.
\newblock Elucidating the design space of diffusion-based generative models, 2022.
\newblock URL \url{https://arxiv.org/abs/2206.00364}.

\bibitem[Kim et~al.(2025)Kim, Yoon, et~al.]{kim2025flowits}
Jaihoon Kim, Taehoon Yoon, et~al.
\newblock Inference-time scaling for flow models via stochastic generation and rollover budget forcing.
\newblock \emph{arXiv:2503.19385}, 2025.

\bibitem[Labs et~al.(2025)Labs, Batifol, Blattmann, Boesel, Consul, Diagne, Dockhorn, English, English, Esser, Kulal, Lacey, Levi, Li, Lorenz, Müller, Podell, Rombach, Saini, Sauer, and Smith]{flux}
Black~Forest Labs, Stephen Batifol, Andreas Blattmann, Frederic Boesel, Saksham Consul, Cyril Diagne, Tim Dockhorn, Jack English, Zion English, Patrick Esser, Sumith Kulal, Kyle Lacey, Yam Levi, Cheng Li, Dominik Lorenz, Jonas Müller, Dustin Podell, Robin Rombach, Harry Saini, Axel Sauer, and Luke Smith.
\newblock Flux.1 kontext: Flow matching for in-context image generation and editing in latent space, 2025.
\newblock URL \url{https://arxiv.org/abs/2506.15742}.

\bibitem[Lightman et~al.(2023)]{lightman2023verify}
Hagai Lightman et~al.
\newblock Let's verify step by step.
\newblock \emph{arXiv:2305.20050}, 2023.

\bibitem[Lin et~al.(2023)Lin, Akin, Rao, Hie, Zhu, Lu, Smetanin, Verkuil, Kabeli, Shmueli, dos Santos~Costa, Fazel-Zarandi, Sercu, Candido, and Rives]{esmfold}
Zeming Lin, Halil Akin, Roshan Rao, Brian Hie, Zhongkai Zhu, Wenting Lu, Nikita Smetanin, Robert Verkuil, Ori Kabeli, Yaniv Shmueli, Allan dos Santos~Costa, Maryam Fazel-Zarandi, Tom Sercu, Salvatore Candido, and Alexander Rives.
\newblock Evolutionary-scale prediction of atomic-level protein structure with a language model.
\newblock \emph{Science}, 379\penalty0 (6637):\penalty0 1123--1130, 2023.
\newblock \doi{10.1126/science.ade2574}.
\newblock URL \url{https://www.science.org/doi/abs/10.1126/science.ade2574}.

\bibitem[Lipman et~al.(2023)]{lipman2023fm}
Yaron Lipman et~al.
\newblock Flow matching for generative modeling.
\newblock \emph{arXiv:2210.02747}, 2023.

\bibitem[Ma et~al.(2024)]{ma2024sit}
Nanye Ma et~al.
\newblock {SiT}: Exploring flow and diffusion-based generative models with scalable interpolant transformers.
\newblock \emph{arXiv:2401.08740}, 2024.

\bibitem[Ma et~al.(2025)]{ma2025diffits}
Nanye Ma et~al.
\newblock Inference-time scaling for diffusion models beyond scaling denoising steps.
\newblock \emph{arXiv:2501.09732}, 2025.

\bibitem[Morshed \& Boddeti(2025)Morshed and Boddeti]{morshed2025diverseflowsampleefficientdiversemode}
Mashrur~M. Morshed and Vishnu Boddeti.
\newblock Diverseflow: Sample-efficient diverse mode coverage in flows, 2025.
\newblock URL \url{https://arxiv.org/abs/2504.07894}.

\bibitem[Nam et~al.(2025)Nam, Liu, Winter, Jun, Yang, and Gómez-Bombarelli]{nam2025flowmatchingacceleratedsimulation}
Juno Nam, Sulin Liu, Gavin Winter, KyuJung Jun, Soojung Yang, and Rafael Gómez-Bombarelli.
\newblock Flow matching for accelerated simulation of atomic transport in materials, 2025.
\newblock URL \url{https://arxiv.org/abs/2410.01464}.

\bibitem[{OpenAI}(2024)]{openai2024o1}
{OpenAI}.
\newblock The \textsc{OpenAI} o1 technical report.
\newblock \url{https://openai.com/research/o1}, 2024.
\newblock accessed May 2025.

\bibitem[Oquab et~al.(2024)Oquab, Darcet, Moutakanni, Vo, Szafraniec, Khalidov, Fernandez, Haziza, Massa, El-Nouby, Assran, Ballas, Galuba, Howes, Huang, Li, Misra, Rabbat, Sharma, Synnaeve, Xu, Jegou, Mairal, Labatut, Joulin, and Bojanowski]{oquab2024dinov2learningrobustvisual}
Maxime Oquab, Timothée Darcet, Théo Moutakanni, Huy Vo, Marc Szafraniec, Vasil Khalidov, Pierre Fernandez, Daniel Haziza, Francisco Massa, Alaaeldin El-Nouby, Mahmoud Assran, Nicolas Ballas, Wojciech Galuba, Russell Howes, Po-Yao Huang, Shang-Wen Li, Ishan Misra, Michael Rabbat, Vasu Sharma, Gabriel Synnaeve, Hu~Xu, Hervé Jegou, Julien Mairal, Patrick Labatut, Armand Joulin, and Piotr Bojanowski.
\newblock Dinov2: Learning robust visual features without supervision, 2024.
\newblock URL \url{https://arxiv.org/abs/2304.07193}.

\bibitem[Salimans et~al.(2016)Salimans, Goodfellow, Zaremba, Cheung, Radford, and Chen]{salimans2016inception}
Tim Salimans, Ian Goodfellow, Wojciech Zaremba, Vicki Cheung, Alec Radford, and Xi~Chen.
\newblock Improved techniques for training gans, 2016.
\newblock URL \url{https://arxiv.org/abs/1606.03498}.

\bibitem[Song \& Ermon(2021)Song and Ermon]{song2021score}
Yang Song and Stefano Ermon.
\newblock Score-based generative modeling through stochastic differential equations.
\newblock \emph{ICLR}, 2021.

\bibitem[Szegedy et~al.(2015)Szegedy, Vanhoucke, Ioffe, Shlens, and Wojna]{szegedy2015rethinkinginceptionarchitecturecomputer}
Christian Szegedy, Vincent Vanhoucke, Sergey Ioffe, Jonathon Shlens, and Zbigniew Wojna.
\newblock Rethinking the inception architecture for computer vision, 2015.
\newblock URL \url{https://arxiv.org/abs/1512.00567}.

\bibitem[Tong et~al.(2024)Tong, Fatras, Malkin, et~al.]{tong2024otfm}
A.~Tong, K.~Fatras, N.~Malkin, et~al.
\newblock Improving and generalizing flow-based generative models with minibatch optimal transport.
\newblock \emph{Transactions on Machine Learning Research}, 2024.

\end{thebibliography}
\bibliographystyle{iclr2026_conference}

\appendix

\section{Inference Algorithm Implementations}
\label{sec:algorithms}

This section provides detailed algorithmic descriptions for the inference methods introduced in Section 4.3.

\subsection{Random Search (Best-of-$N$)}
\label{app:random-search}

\begin{algorithm}[H]
\caption{Random Search (Best-of-$N$)}
\label{alg:best-of-n}
\begin{algorithmic}[1]
\Require Flow matching model $v_\theta$, verifier function $r(\cdot)$, number of samples $N$, number to retain $K$
\Ensure Top-$K$ generated samples
\For{$i = 1$ to $N$}
    \State $x_0^{(i)} \sim \pi_{\text{ref}}$ \Comment{Sample initial condition}
    \State $x_1^{(i)} \leftarrow \text{ODESolve}(v_\theta, x_0^{(i)})$ \Comment{Deterministic sampling}
    \State $s^{(i)} \leftarrow r(x_1^{(i)})$ \Comment{Compute verifier score}
\EndFor
\State \Return Top-$K$ samples by score $\{s^{(i)}\}$
\end{algorithmic}
\end{algorithm}

\subsection{Noise Search}
\label{app:noise-search}

\begin{algorithm}[H]
\caption{Multi-Round Noise Search}
\label{alg:noise-search}
\begin{algorithmic}[1]
\Require Flow matching model $v_\theta$, initial noise $x_0$, noise injection method $\mathcal{N}(\cdot)$, verifier function $r(\cdot)$, rounds $R$, samples per round $N$, candidates per round $K$
\Ensure Generated sample $x_1$
\State candidates $\leftarrow \{x_0\}$ \Comment{Initialize with single candidate}
\For{round $i = 1$ to $R$}
    \State $t_{\text{start}} \leftarrow (i-1) / R$ \Comment{Starting timestep for round}
    \State round\_samples $\leftarrow \{\}$
    \For{each candidate $x_{t_{\text{start}}}$ in candidates}
        \For{$j = 1$ to $N$}
            \State $x_1^{(j)} \leftarrow \text{NoisySample}(v_\theta, x_{t_{\text{start}}}, t_{\text{start}}, \mathcal{N})$
            \State $s^{(j)} \leftarrow r(x_1^{(j)})$
            \State Add $(x_{t_{\text{start}}}, x_1^{(j)}, s^{(j)})$ to round\_samples
        \EndFor
    \EndFor
    \If{$i < R$}
        \State candidates $\leftarrow$ Top-$K$ starting points by final score
    \Else
        \State \Return Best final sample by verifier score
    \EndIf
\EndFor
\end{algorithmic}
\end{algorithm}

\newpage

\subsection{RS + Noise Search}
\label{app:rs-noise-search}

\begin{algorithm}[H]
\caption{Two-Stage RS + Noise Search}
\label{alg:best-of-n-noise-search}
\begin{algorithmic}[1]
\Require Flow matching model $v_\theta$, verifier function $r(\cdot)$, stage 1 samples $N$, top candidates $K$, noise injection method $\mathcal{N}(\cdot)$, search rounds $R$
\Ensure Top-$K$ generated samples
\State \textbf{Stage 1:} $\{x_0^*\} \leftarrow$ BestOfN($v_\theta$, $r$, $N$, $K$) \Comment{Algorithm~\ref{alg:best-of-n}}
\State \textbf{Stage 2:} results $\leftarrow \{\}$
\For{each $x_0^{(i)} \in \{x_0^*\}$}
    \State $x_1^{(i)} \leftarrow$ NoiseSearch($v_\theta$, $x_0^{(i)}$, $\mathcal{N}$, $r$, $R$) \Comment{Algorithm~\ref{alg:noise-search}}
    \State Add $x_1^{(i)}$ to results
\EndFor
\State \Return Top-$K$ samples from results by verifier score
\end{algorithmic}
\end{algorithm}

\section{Additional Experimental Results}
\label{app:additional-results}

This appendix contains supplementary figures and detailed results that support the main findings presented in the paper.

\subsection{Noise and Diversity Study Results}
\label{app:noise-diversity-results}

This section consolidates all experimental results related to noise injection and diversity analysis, including Pareto frontier analysis and individual noise curves.

\begin{figure}[H]
  \centering
  \includegraphics[width=0.8\textwidth]{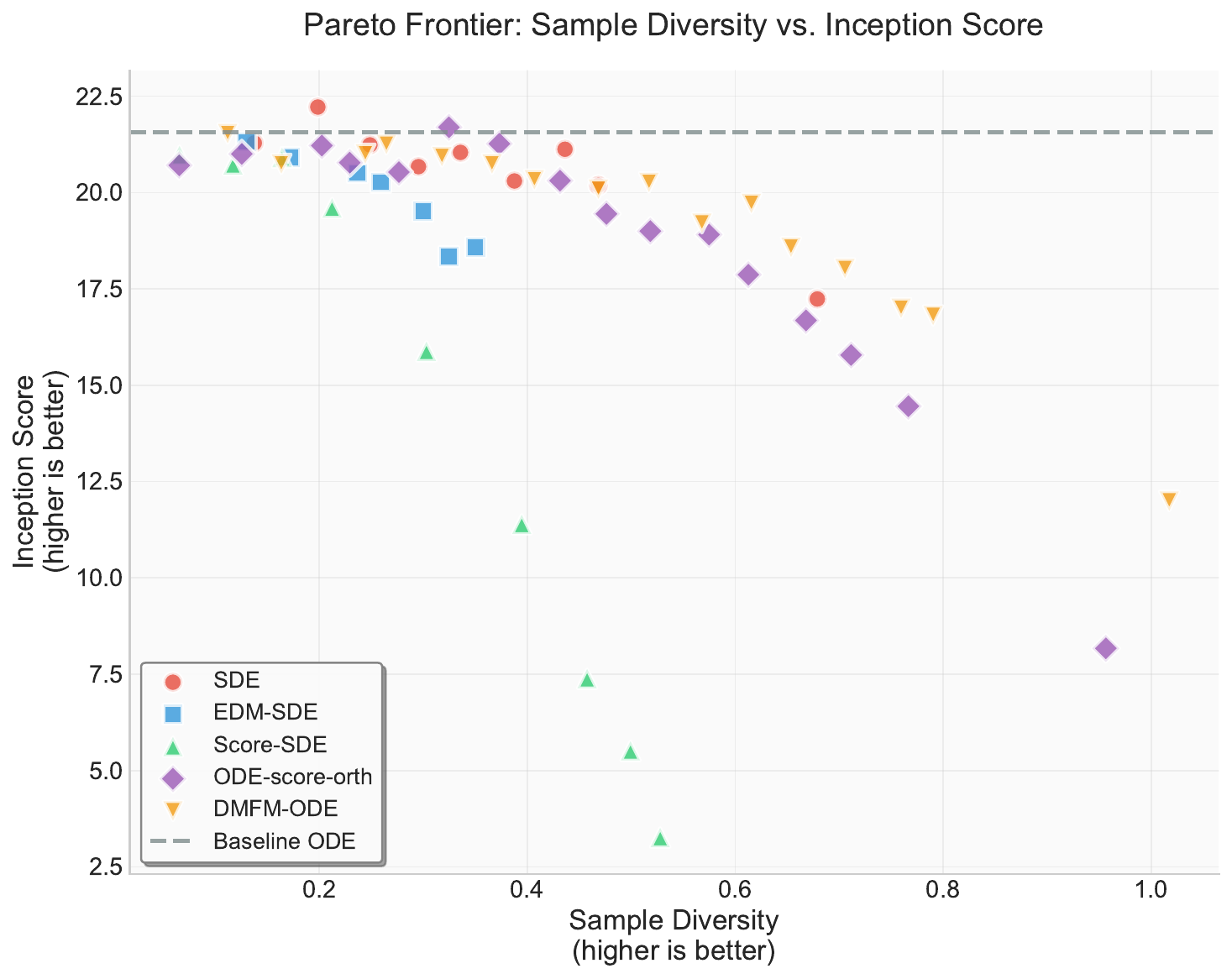}
  \caption{Pareto frontier: sample diversity vs. Inception Score (higher is better on both axes). This complements the FID-based analysis shown in the main text, demonstrating consistent frontier expansion across different quality metrics.}
  \label{fig:pareto-inception}
\end{figure}

\newpage

\begin{figure}[H]
  \centering
  \includegraphics[width=0.7\linewidth]{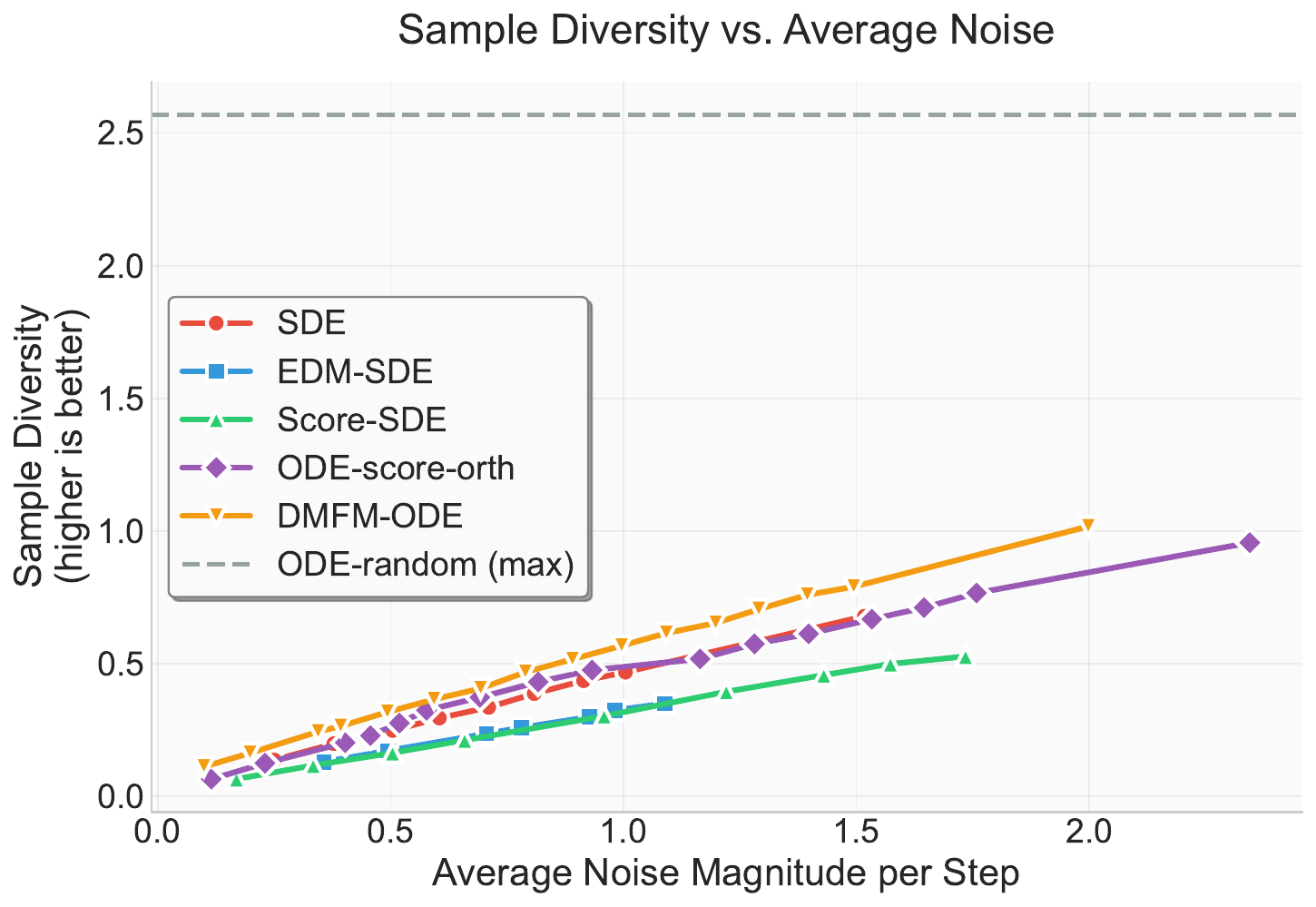}
  \caption{Sample diversity across increasing average noise magnitudes. Higher is better. These curves complement the Pareto analysis by showing per-noise behavior for each method, including the composite DMFM-ODE and baselines.}
  \label{fig:appendix-div-noise}
\end{figure}

\begin{figure}[H]
  \centering
  \begin{minipage}{0.48\textwidth}
    \centering
    \includegraphics[width=\textwidth]{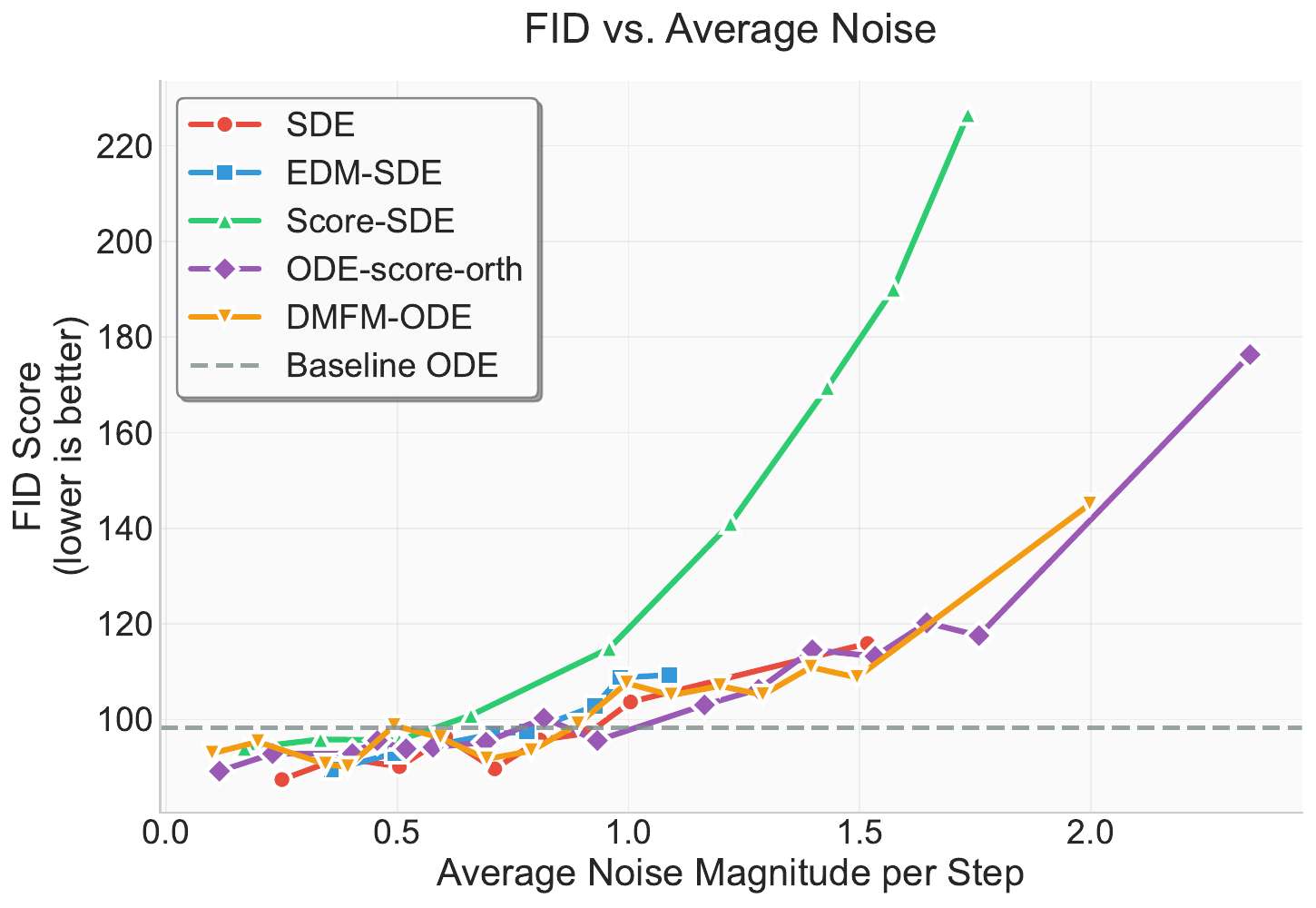}
  \end{minipage}
  \hfill
  \begin{minipage}{0.48\textwidth}
    \centering
    \includegraphics[width=\textwidth]{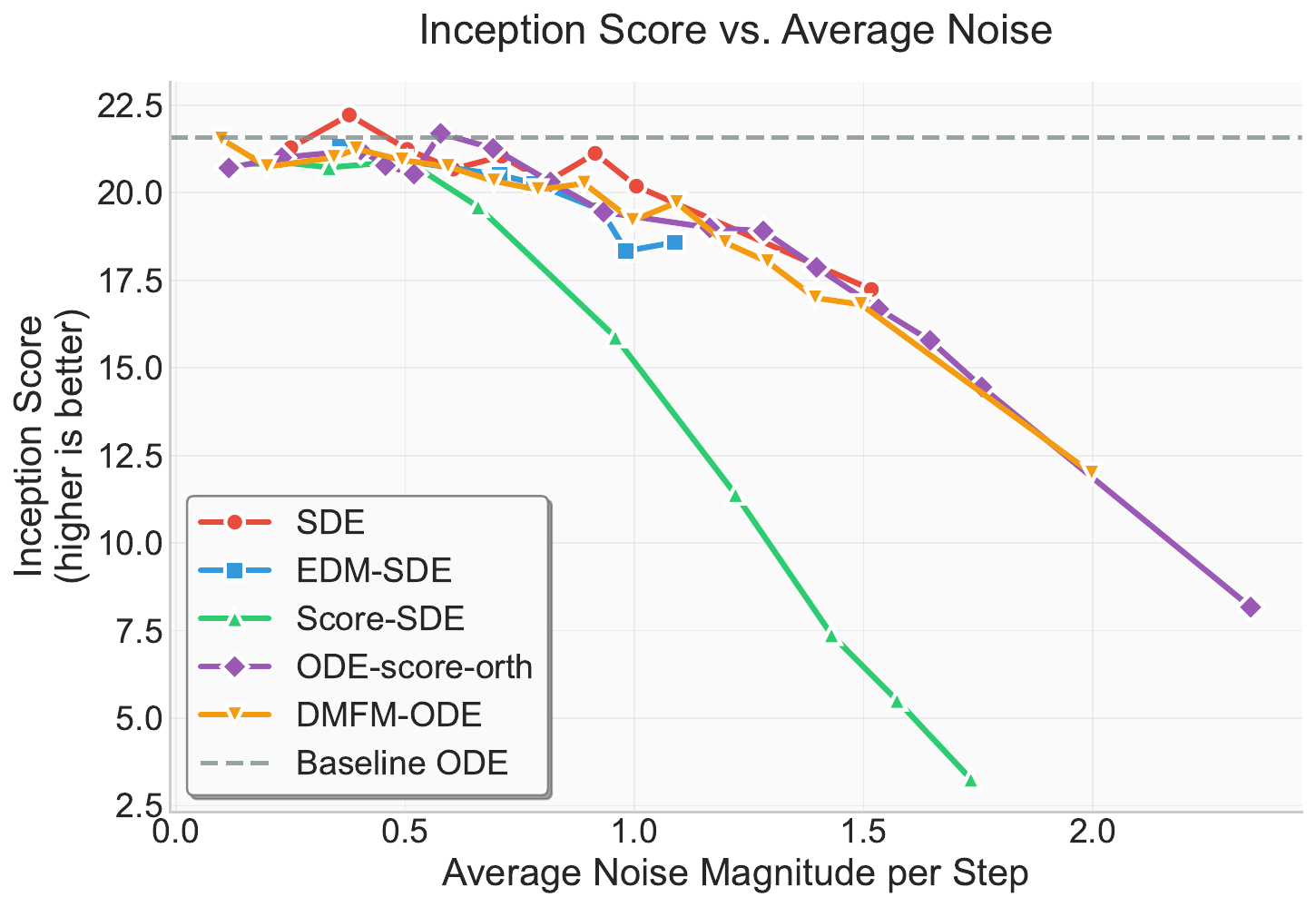}
  \end{minipage}
  \caption{Quality metrics across increasing average noise magnitudes. Left: FID (lower is better). Right: Inception Score (higher is better). These curves complement the Pareto analysis and highlight the robustness of DMFM-ODE compared to SDE baselines as noise increases.}
  \label{fig:appendix-fid-is-noise}
\end{figure}

\newpage

\subsection{Complete Inference-Time Scaling Results}
\label{app:complete-scaling-results}

This section provides complete results for all metrics across both Inception Score-guided and DINO-guided scaling experiments.

\begin{figure}[H]
  \centering
  \begin{minipage}{0.48\textwidth}
    \centering
    \includegraphics[width=\textwidth]{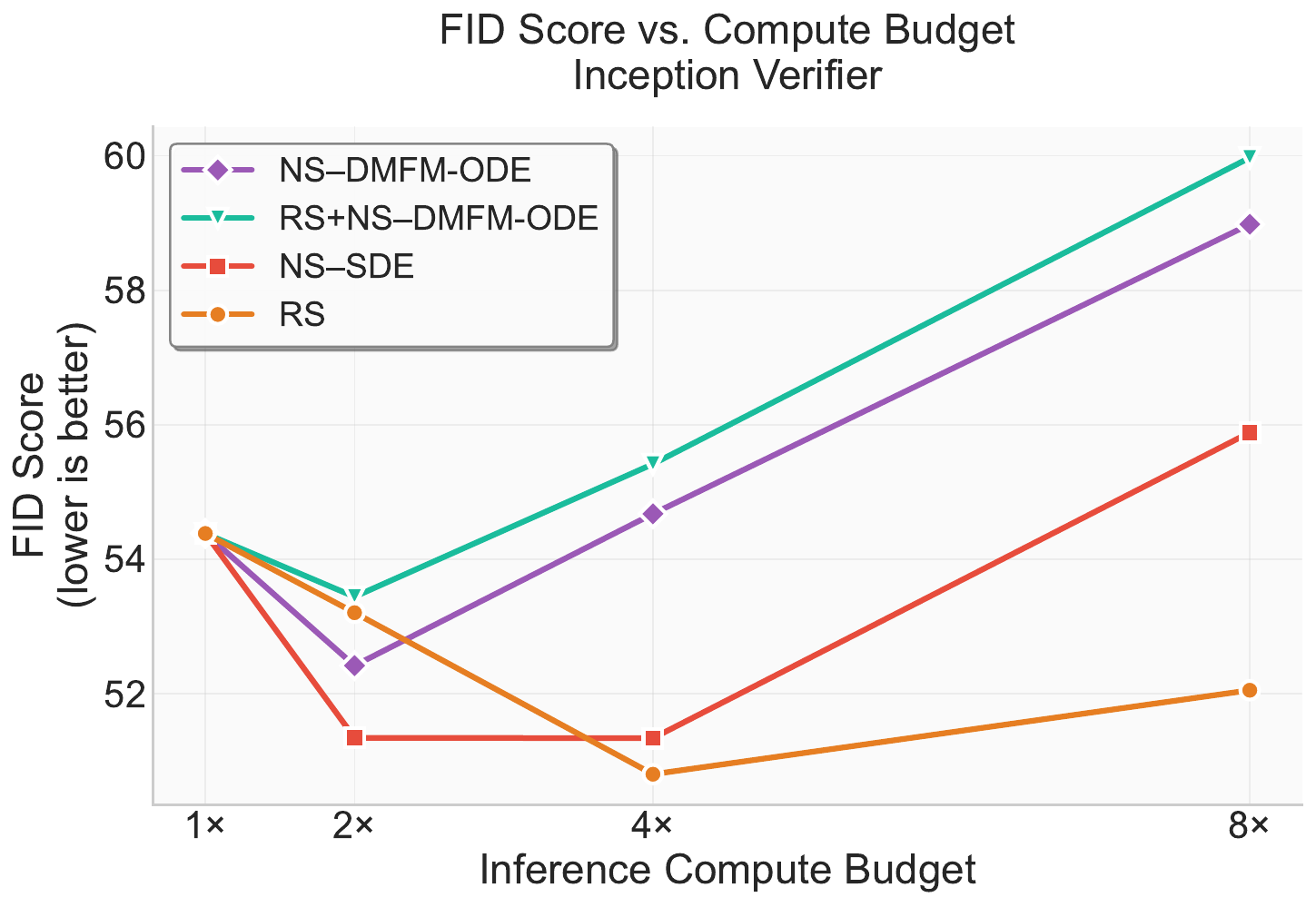}
  \end{minipage}
  \hfill
  \begin{minipage}{0.48\textwidth}
    \centering
    \includegraphics[width=\textwidth]{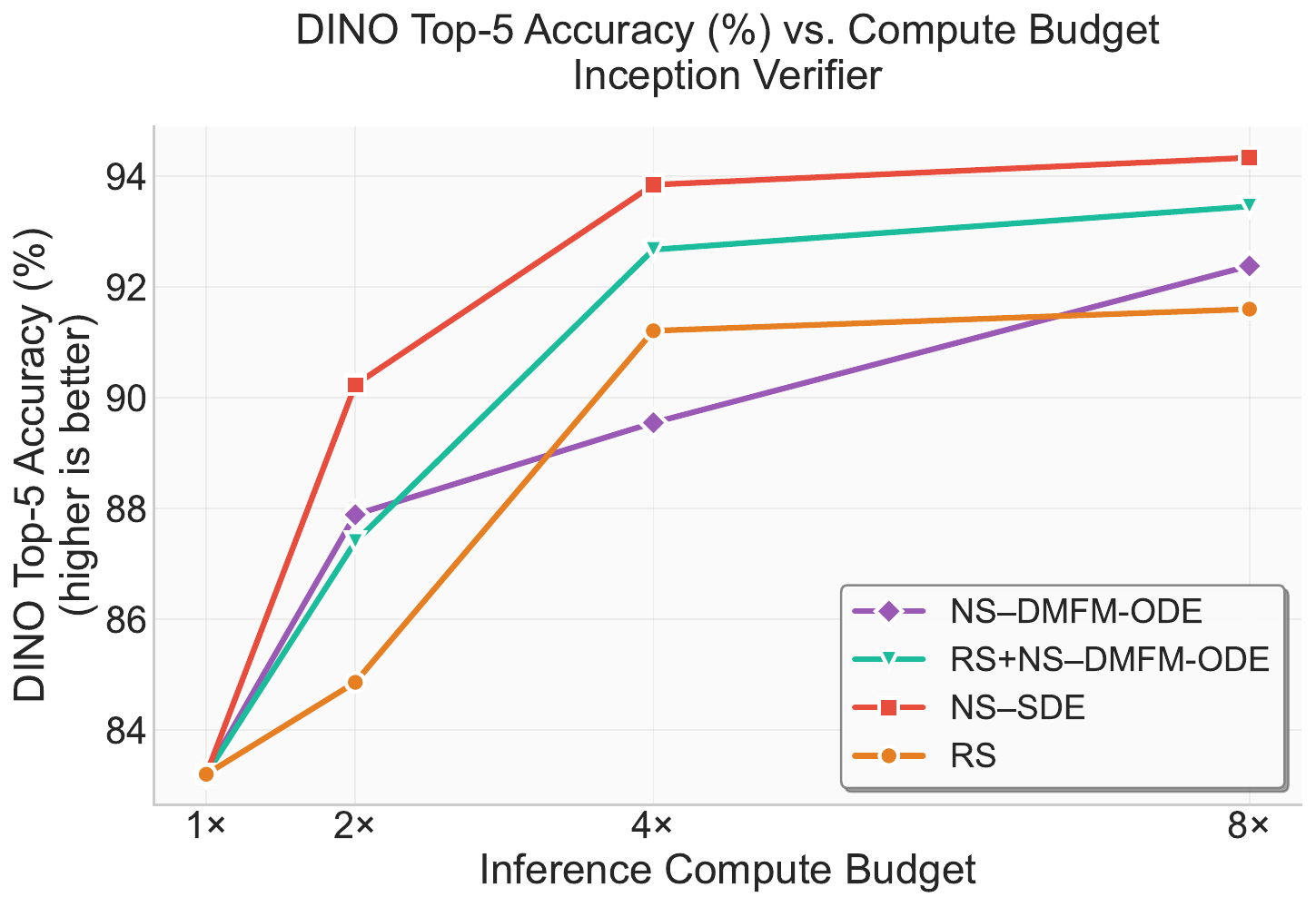}
  \end{minipage}
  \caption{Additional Inception Score-guided scaling results. Left: FID vs. compute budget. Right: DINO Top-5 accuracy vs. compute budget.}
  \label{fig:inception-scaling-complete}
\end{figure}

\begin{figure}[H]
  \centering
  \includegraphics[width=0.48\textwidth]{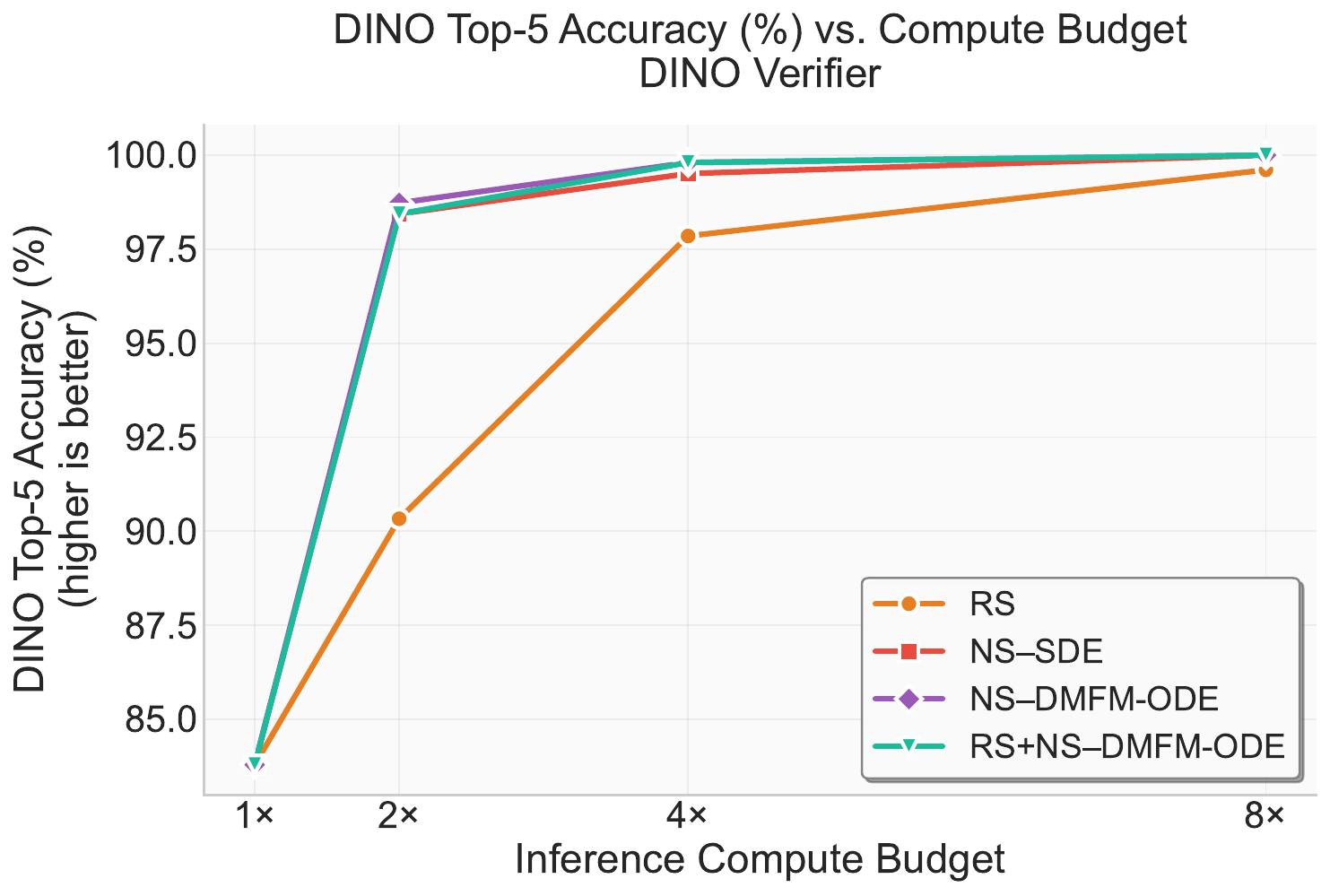}
  \caption{DINO Top-5 accuracy vs. compute budget for DINO-guided scaling experiments.}
  \label{fig:dino-scaling-complete}
\end{figure}

\newpage

\subsection{Geometric Scoring Experiments}
\label{app:geometric-scoring}

We conducted the same protein generation experiments using a geometric scoring function instead of TM-score for sample selection. All experimental parameters remain identical: 64 protein samples of length 100 residues per configuration, compute budgets of 1×, 2×, 4×, and 8×, with the same four inference-time scaling methods.

\begin{figure}[H]
  \centering
  \begin{minipage}{0.48\textwidth}
    \centering
    \includegraphics[width=\textwidth]{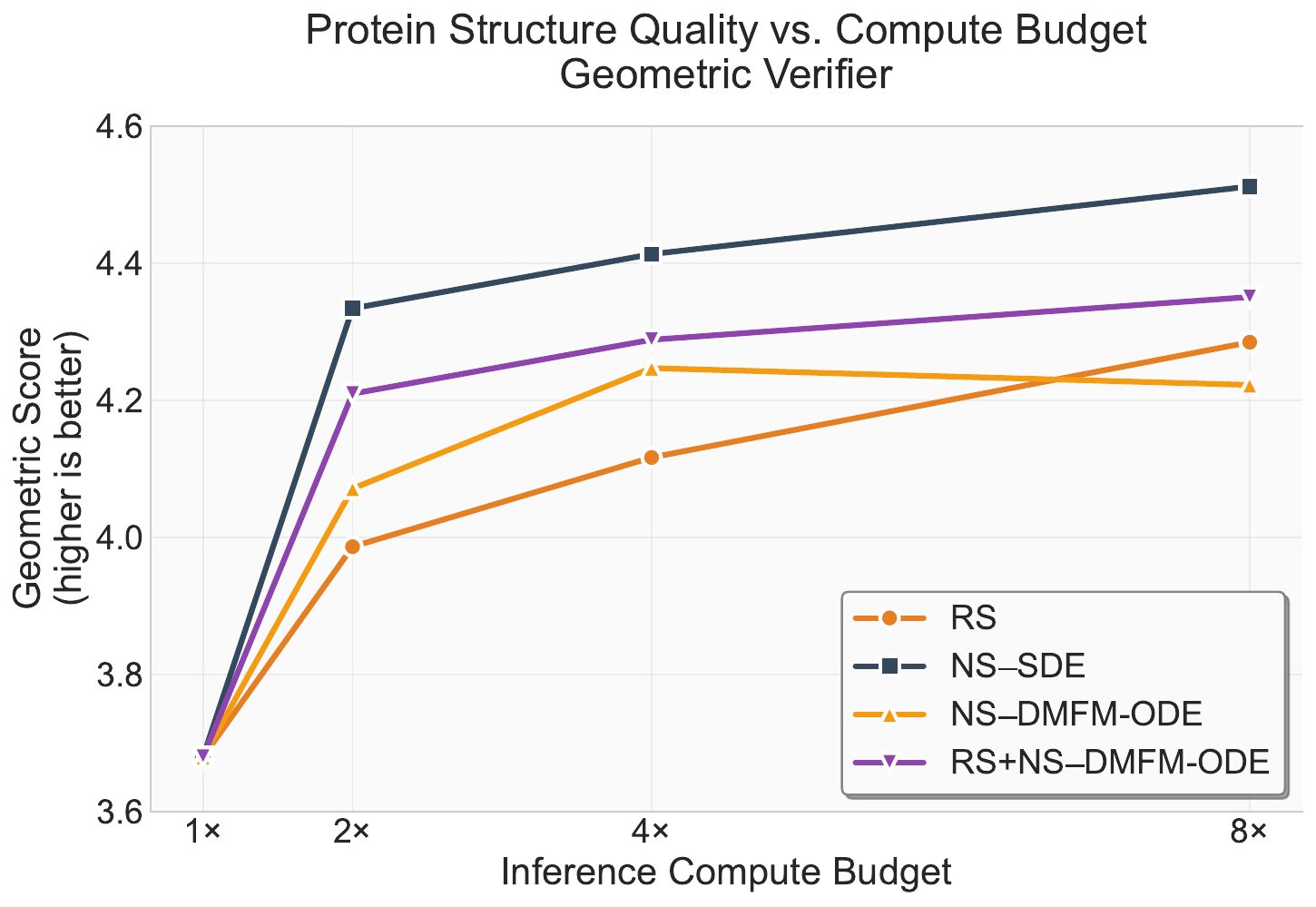}
  \end{minipage}
  \hfill
  \begin{minipage}{0.48\textwidth}
    \centering
    \includegraphics[width=\textwidth]{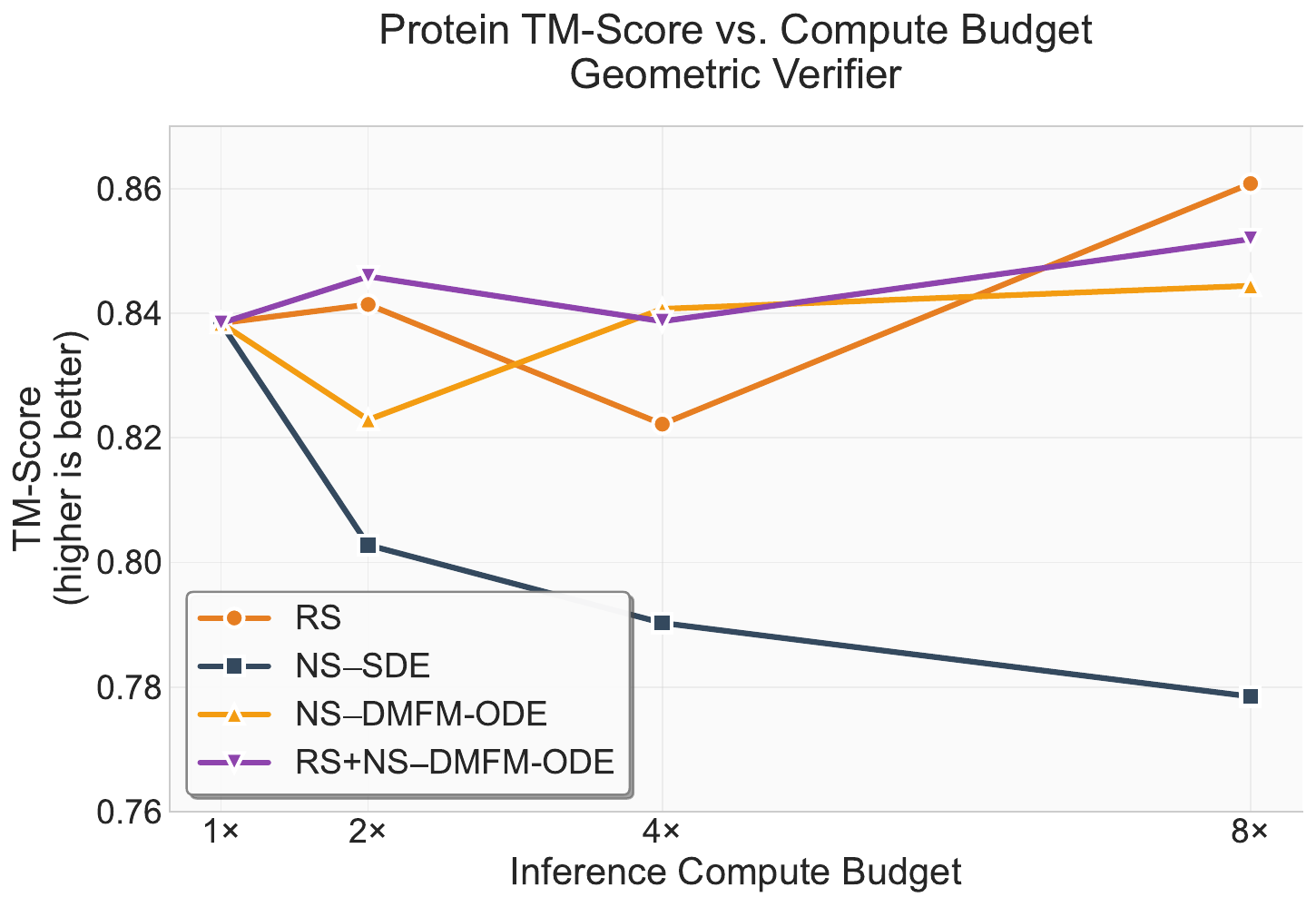}
  \end{minipage}
  \caption{Protein generation results using geometric scoring function. Left: Geometric score vs. compute budget (higher is better). Right: TM-score vs. compute budget, showing how geometric-guided selection affects TM-score performance.}
  \label{fig:geometric-scaling-tm}
\end{figure}

\begin{figure}[H]
  \centering
  \begin{minipage}{0.48\textwidth}
    \centering
    \includegraphics[width=\textwidth]{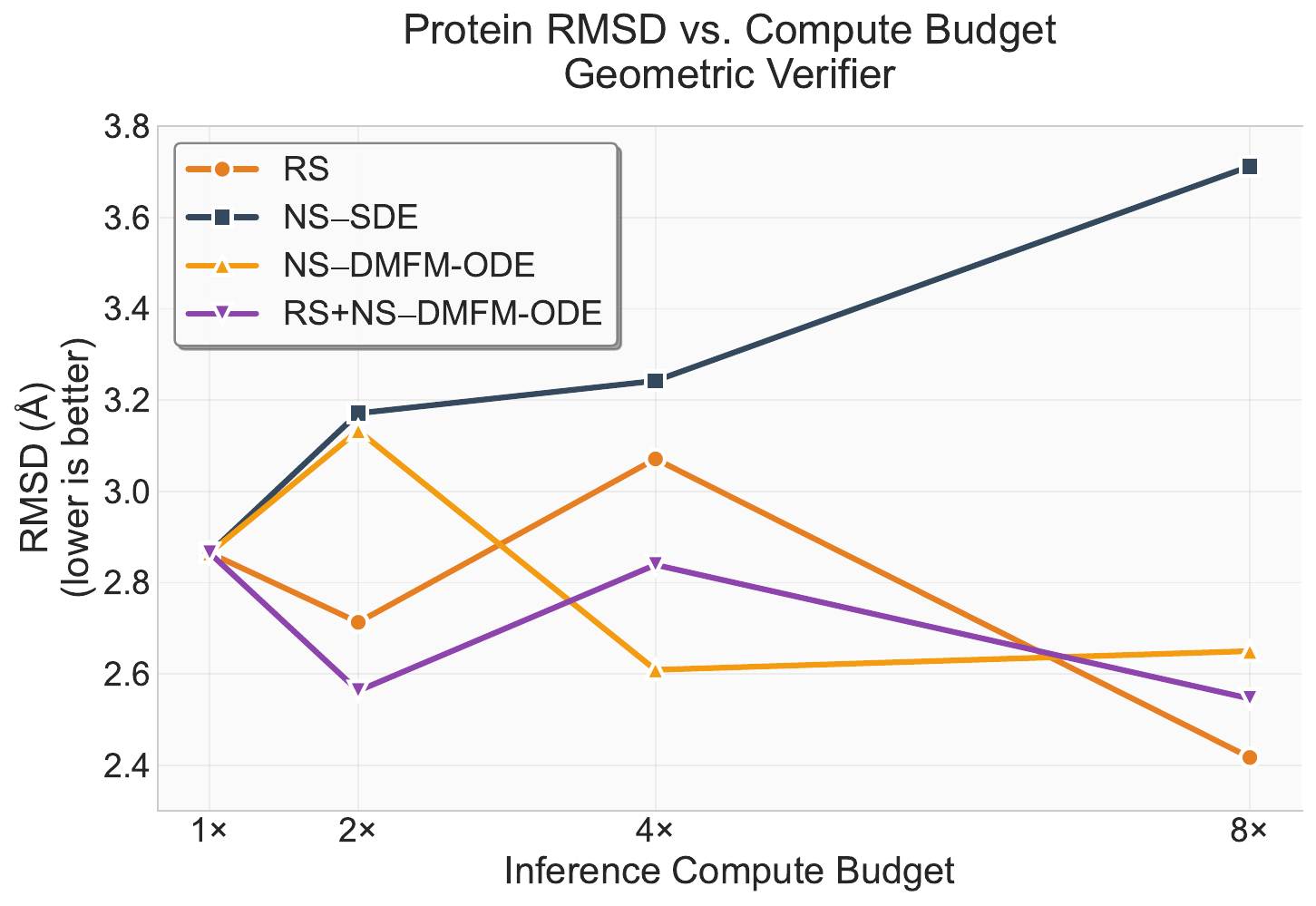}
  \end{minipage}
  \hfill
  \begin{minipage}{0.48\textwidth}
    \centering
    \includegraphics[width=\textwidth]{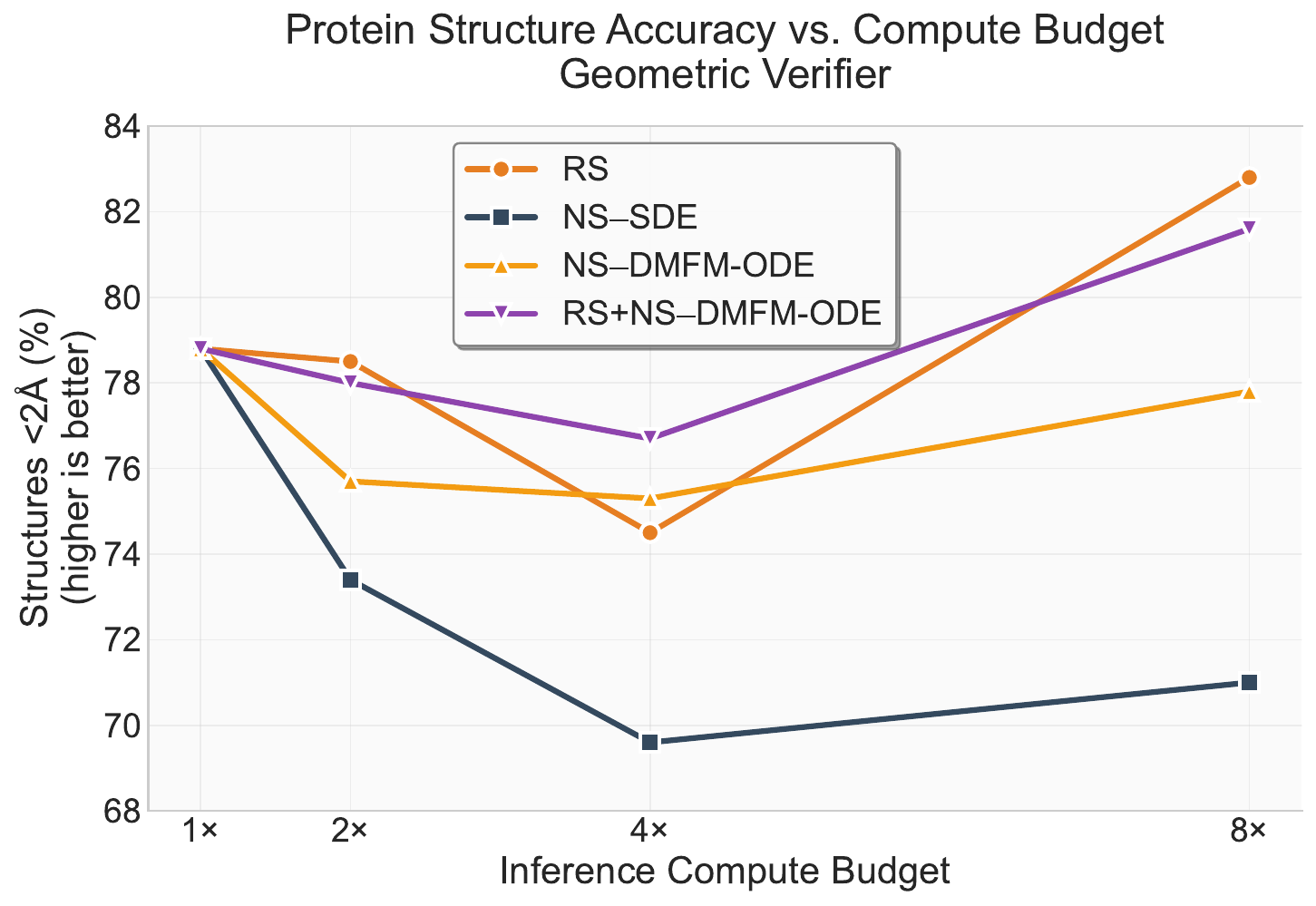}
  \end{minipage}
  \caption{Protein structural quality metrics using geometric scoring function. Left: RMSD vs. compute budget (lower is better). Right: Percentage of structures with RMSD < 2Å vs. compute budget, showing designability improvements.}
  \label{fig:geometric-rmsd-2a}
\end{figure}

The geometric scoring experiments show that while all methods improve with increased compute budget, the relative performance rankings change compared to TM-score selection. This demonstrates that inference-time scaling benefits are robust across different verifiers, though optimal method selection may vary depending on the specific evaluation criteria used.

\newpage

\section{Additional Visual Examples}
\label{app:additional-visual-examples}

This section provides additional visual examples of inference-time compute scaling for ImageNet 256×256 generation, comparing Random Search and our RS+NS–DMFM-ODE method across different compute budgets.

\begin{figure}[H]
  \centering
  \includegraphics[width=0.9\textwidth]{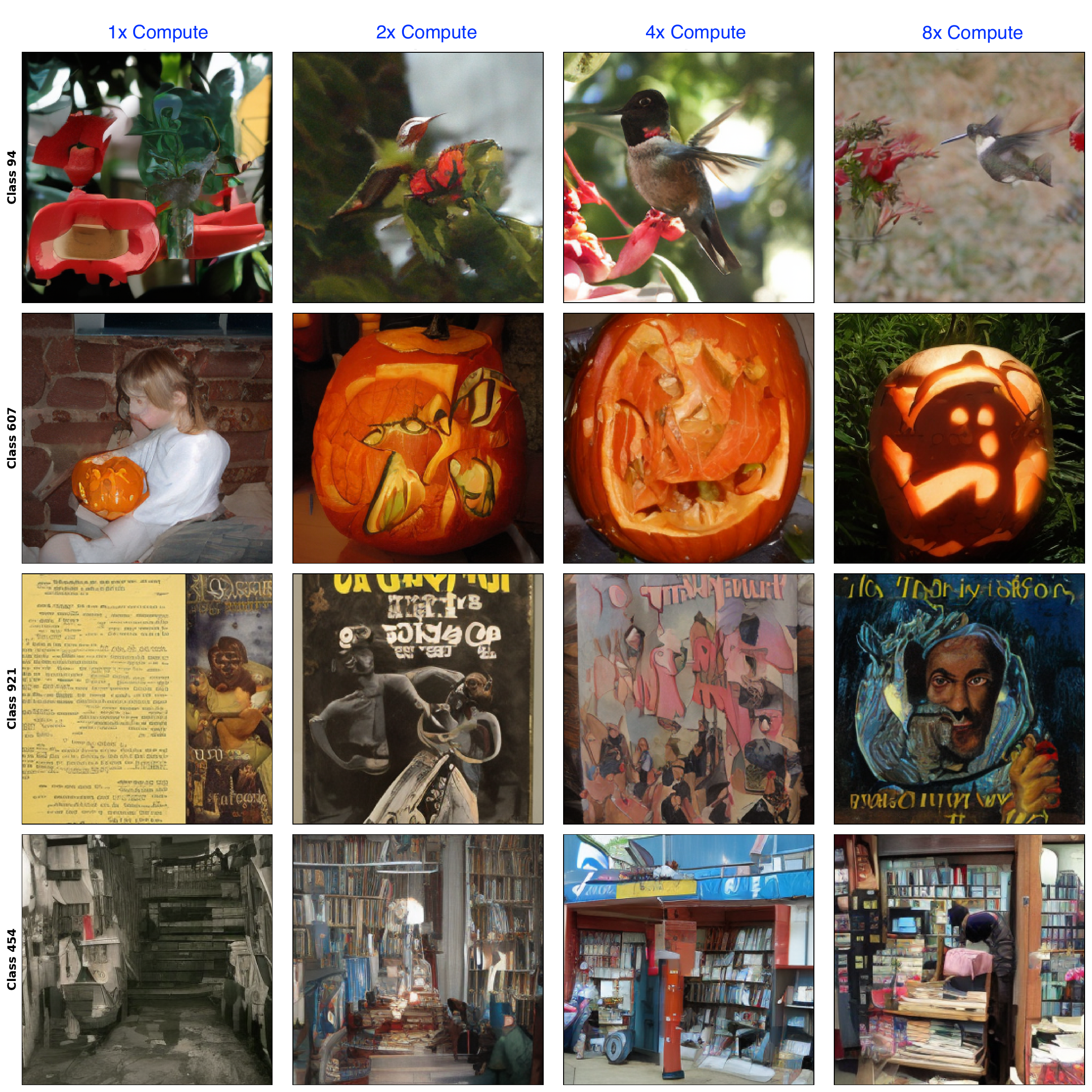}
  \caption{Random Search sampling examples across compute budgets using DINO verifier on ImageNet 256×256.}
  \label{fig:random-search-dino-examples}
\end{figure}

\newpage

\begin{figure}[H]
  \centering
  \includegraphics[width=0.9\textwidth]{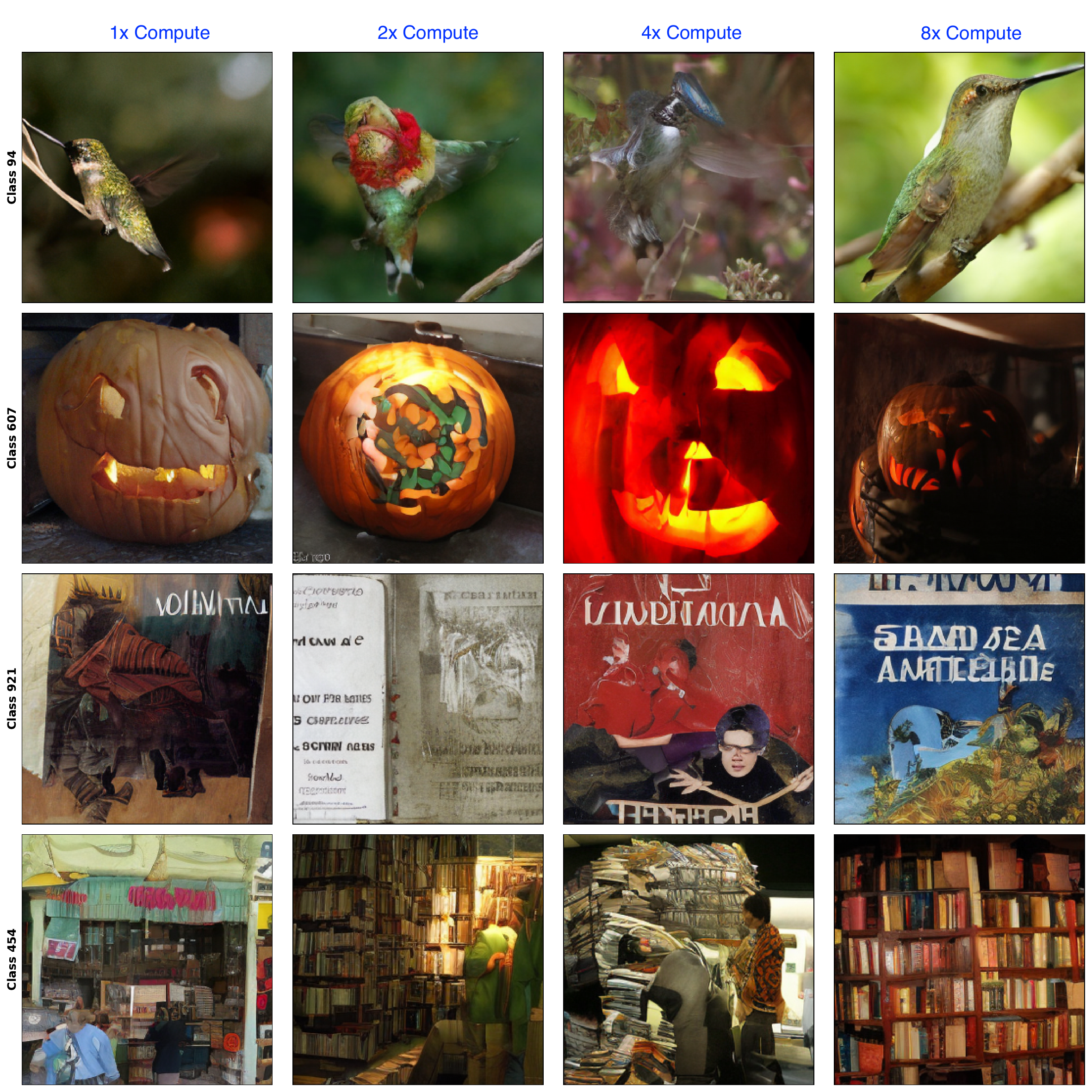}
  \caption{RS+NS–DMFM-ODE sampling examples across compute budgets using DINO verifier on ImageNet 256×256.}
  \label{fig:two-stage-dino-examples}
\end{figure}

\begin{figure}[H]
  \centering
  \includegraphics[width=0.9\textwidth]{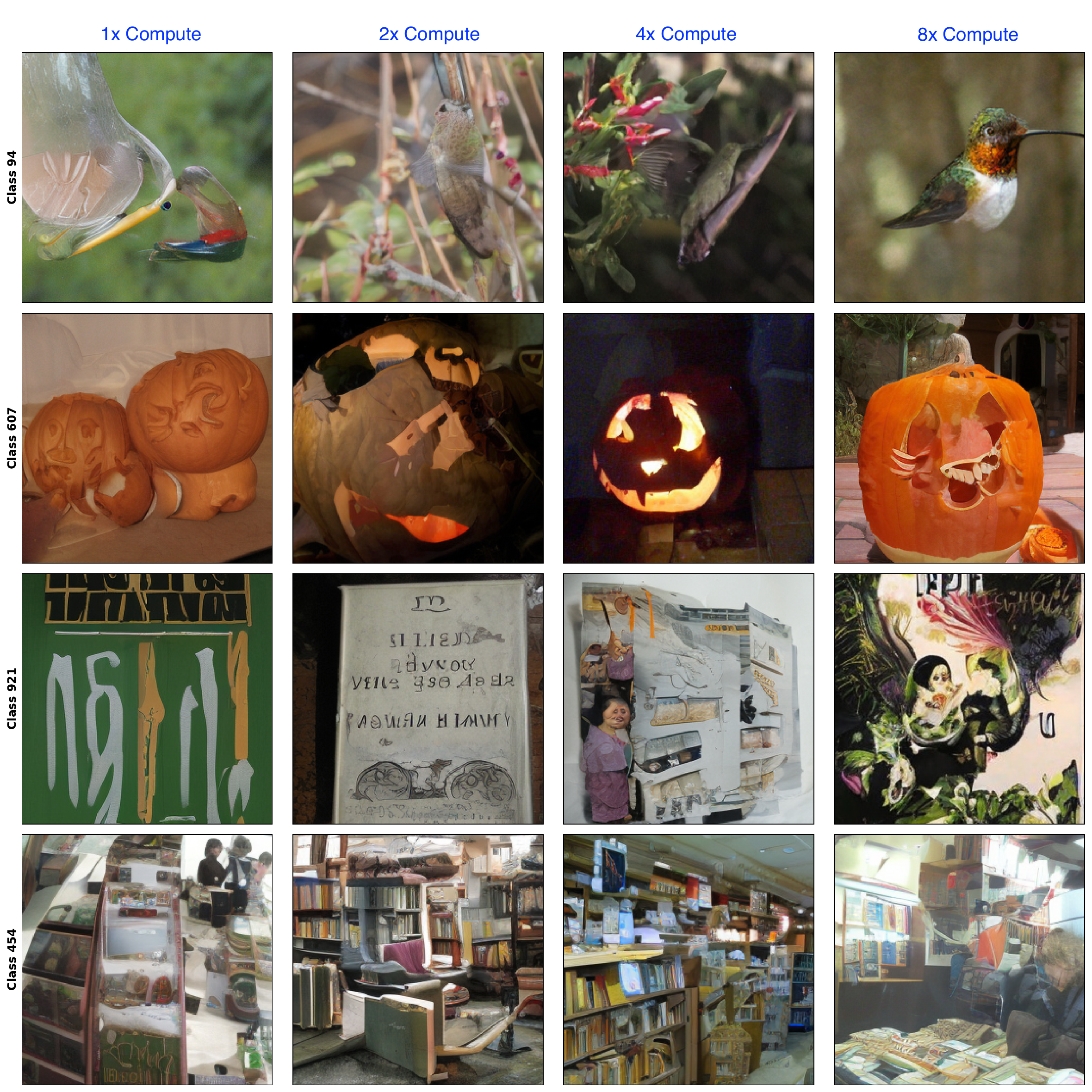}
  \caption{Random Search sampling examples across compute budgets using Inception Score verifier on ImageNet 256×256.}
  \label{fig:random-search-inception-examples}
\end{figure}

\begin{figure}[H]
  \centering
  \includegraphics[width=0.9\textwidth]{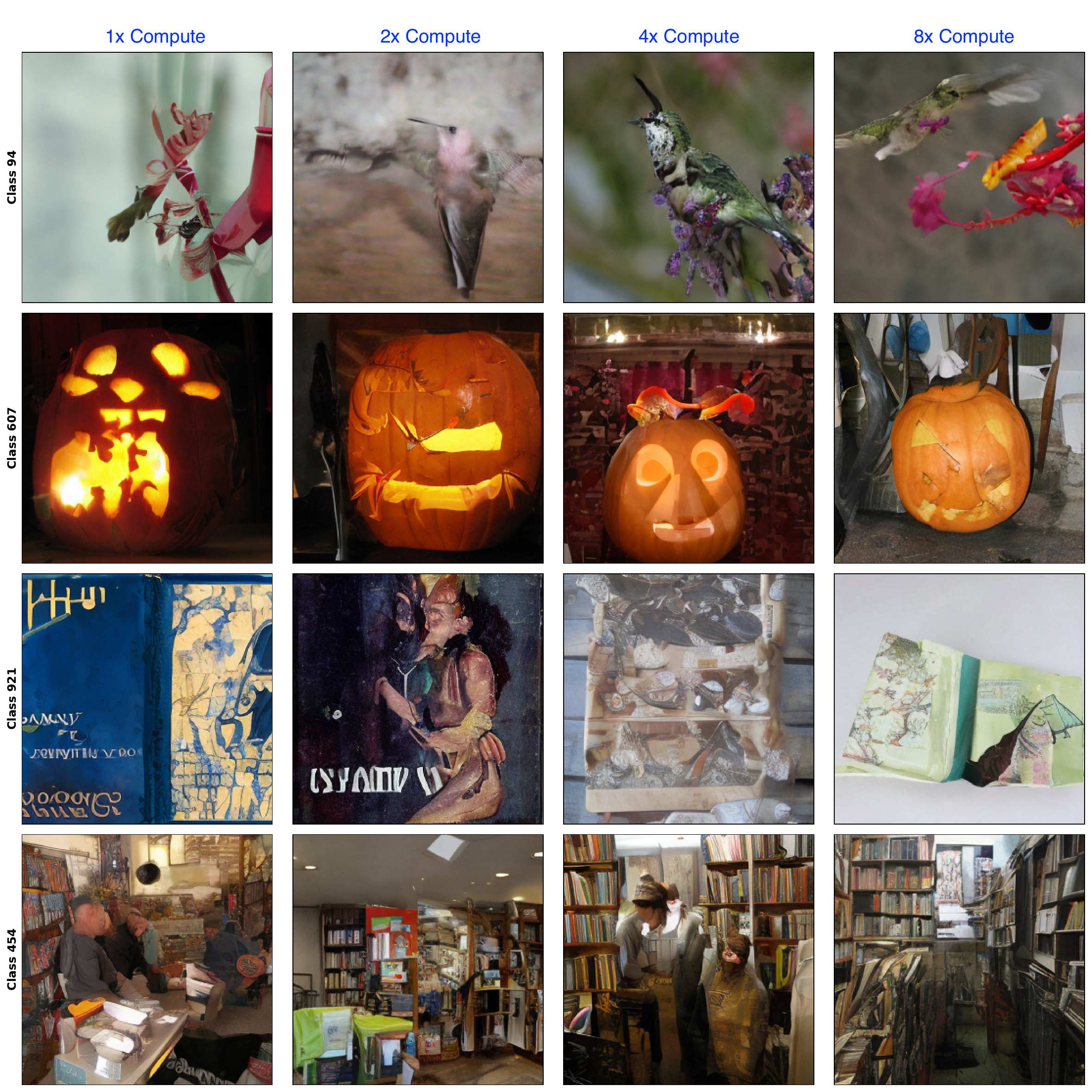}
  \caption{RS+NS–DMFM-ODE sampling examples across compute budgets using Inception Score verifier on ImageNet 256×256.}
  \label{fig:two-stage-inception-examples}
\end{figure}

\newpage

\section{Implementation Details}
\label{app:implementation-details}

\subsection{Experimental Hyperparameters}
\label{app:experimental-hyperparameters}

Table~\ref{tab:hyperparameters} summarizes the key hyperparameters used across all experiments. These values were selected based on preliminary experiments to balance sample quality and computational efficiency.

\begin{table}[H]
\centering
\begin{tabular}{@{}lcc@{}}
\textbf{Parameter} & \textbf{ImageNet} & \textbf{FoldFlow} \\
\midrule
Model & SiT-XL/2 & FoldFlow \\
Number of samples & 1,024 & 64 \\
Integration timesteps & 20 & 50 \\
Protein length & N/A & 100 residues \\
\midrule
\textbf{Score-orthogonal methods} & & \\
Noise scale ($\lambda$) & 0.9 & 0.3 \\
\midrule
\textbf{SDE methods} & & \\
Noise scale ($\sigma$) & 0.14 & 0.2 \\
\midrule
\textbf{Compute budgets} & 1×, 2×, 4×, 8× & 1×, 2×, 4×, 8× \\
\textbf{Verifiers} & Inception, DINO & TM-score \\
\bottomrule
\end{tabular}
\caption{Experimental hyperparameters for ImageNet and FoldFlow experiments.}
\label{tab:hyperparameters}
\end{table}

\subsection{ImageNet Implementation Details}
\label{app:imagenet-details}

For ImageNet experiments, we use the pretrained SiT-XL/2 model with 20 integration timesteps, following the standard configuration for efficient sampling. Noise scales were selected to provide substantial diversity without degrading sample quality, as validated in our noise study (Section~\ref{sec:noise_study}). All ImageNet experiments generate 1,024 samples per configuration to ensure robust statistical evaluation across the four metrics (FID, Inception Score, DINO Top-1, DINO Top-5). Verifier-based selection uses either Inception Score or DINO features, enabling comparison of verifier-method alignment effects. Noise Search uses 9 rounds of noise injection distributed across the trajectory, with round start times at: \texttt{[0.0, 0.2, 0.4, 0.6, 0.75, 0.8, 0.85, 0.9, 0.95]}. This configuration provides systematic exploration of the trajectory space while maintaining computational efficiency.

\subsection{FoldFlow Implementation Details}
\label{app:foldflow-details}

FoldFlow experiments focus on proteins of length 100 residues, providing a manageable complexity for systematic evaluation while representing realistic protein design scenarios. The model uses 50 integration timesteps by default for protein structure generation. Noise scales reflect the different data characteristics and model sensitivities compared to image generation, calibrated to provide meaningful exploration while preserving protein structural validity. The smaller sample size (64 proteins) reflects the computational cost of protein generation and evaluation, while still providing sufficient data for reliable TM-score estimation and method comparison. Noise Search uses the same 9-round configuration as ImageNet experiments, adapted to the protein generation timestep schedule.

\section{Score-Orthogonal Noise Injection}
\label{app:score-orthogonal-analysis}

We present a mathematical analysis of score-orthogonal perturbations and their effect on the continuity equation in flow matching.

\subsection{Score-Orthogonal Perturbation Construction}

Consider perturbations of the form:
\[
w_t(x) = \Pi_{\perp s_t}\,\varepsilon = \bigl(I - \hat s_t \hat s_t^\top\bigr)\,\varepsilon,
\]
where \(\hat s_t = s_t/\|s_t\|\), \(s_t = \nabla_x \log p_t(x)\) is the score function, and \(\varepsilon \sim \mathcal{N}(0,I)\).

\textbf{Theorem 1 (Score Orthogonality).} The perturbation \(w_t(x)\) satisfies \(s_t \cdot w_t = 0\) by construction.

\textbf{Proof.} Direct computation yields:
\[
s_t \cdot w_t = s_t \cdot \bigl(I - \hat s_t \hat s_t^\top\bigr)\,\varepsilon = s_t \cdot \varepsilon - (s_t \cdot \hat s_t)(\hat s_t \cdot \varepsilon) = s_t \cdot \varepsilon - \|s_t\|(\hat s_t \cdot \varepsilon) = 0.
\]

\subsection{Effect on the Continuity Equation}

The continuity equation \(\partial_t p_t + \nabla \cdot (p_t v_t) = 0\) is modified under our perturbed dynamics \(dx_t = (v_t + w_t)dt\) to:
\[
\partial_t p_t + \nabla \cdot (p_t (v_t + w_t)) = 0.
\]

\textbf{Theorem 2 (Probability-Weighted Divergence Reduction).} Score-orthogonal perturbations minimize the probability-weighted divergence contribution:
\[
\nabla \cdot (p_t w_t) = p_t \nabla \cdot w_t.
\]

\textbf{Proof.} Using the product rule:
\[
\nabla \cdot (p_t w_t) = p_t \nabla \cdot w_t + w_t \cdot \nabla p_t = p_t \nabla \cdot w_t + w_t \cdot (p_t s_t).
\]
Since \(w_t \cdot s_t = 0\) by Theorem 1, we obtain \(\nabla \cdot (p_t w_t) = p_t \nabla \cdot w_t\).

\textbf{Remark:} While \(\nabla \cdot w_t \neq 0\) in general due to the \(x\)-dependence of the projection matrix, the score orthogonality eliminates the dominant \(p_t s_t\) contribution to density evolution, providing a principled approach to noise injection that minimizes violations of the continuity equation while maintaining sufficient stochasticity for effective search.

\end{document}